\documentclass{article}
\usepackage{nips15submit_e,times}
\usepackage{hyperref}
\usepackage{url}
\usepackage{caption}
\usepackage{subcaption}

\usepackage{amsmath}
\usepackage{amsfonts}
\usepackage{amssymb}
\usepackage{graphicx}
\usepackage{bm}
\usepackage{tabularx}
\usepackage{ragged2e}
\newcolumntype{Y}{>{\RaggedRight\arraybackslash}X}

\usepackage{booktabs}
\usepackage{paralist}
\usepackage[leftcaption]{sidecap}
\sidecaptionvpos{figure}{t}

\usepackage{ifthen}
\newcommand{\todo}[2][]{
  \ifthenelse { \equal{#1}{} }
  {{\color{red}(#2)}}
  {{\color{red}\textbf{TODO(#1):} #2}}
}

\usepackage{textcomp}

\newcommand{\igate}{i}
\newcommand{\fgate}{f}
\newcommand{\wtmat}[2]{W_{#1 #2}}
\newcommand{\ogate}{o}
\newcommand{\state}{c}

\nipsfinalcopy

\title{Teaching Machines to Read and Comprehend}

\author{
  Karl Moritz Hermann$^\dag$ \quad Tom\'a\v{s} Ko\v{c}isk\'y$^{\dag\ddag}$ \quad Edward Grefenstette$^\dag$ \\
  {\bf Lasse Espeholt$^\dag$ \quad Will Kay$^\dag$ \quad Mustafa Suleyman$^\dag$
  \quad Phil Blunsom$^{\dag\ddag}$} \\
$^\dag$Google DeepMind \quad
$^\ddag$University of Oxford \\
\texttt{\{kmh,tkocisky,etg,lespeholt,wkay,mustafasul,pblunsom\}@google.com} \\
}

\begin{document}

\maketitle

\begin{abstract}
  Teaching machines to read natural language documents remains an elusive
  challenge. Machine reading systems can be tested on their ability to answer
  questions posed on the contents of documents that they have seen, but
  until now large scale training and test datasets have been missing for this
  type of evaluation.
  In this work we define a new methodology that resolves this bottleneck
  and provides large scale supervised reading comprehension data.
  This allows us to develop a class of attention based
  deep neural networks that learn to read real documents
  and answer complex questions with minimal prior knowledge of language
  structure.
\end{abstract}

\section{Introduction}
\label{introduction}

Progress on the path from shallow bag-of-words information retrieval
algorithms to machines capable of reading and understanding documents has been
slow. Traditional approaches to machine reading and comprehension have been
based on either hand engineered grammars \cite{Riloff:2000:RQA}, or information
extraction methods of detecting predicate argument triples that can later be
queried as a relational database \cite{Poon:2010:MRU}.
Supervised machine learning approaches have largely been absent from this space
due to both the lack of large scale training datasets, and the difficulty in
structuring statistical models flexible enough to learn to exploit document
structure.

While obtaining supervised natural language reading comprehension data has
proved difficult, some researchers have explored generating synthetic narratives
and queries \cite{Weston:2014:MN,Sukhbaatar:2015}. Such approaches allow
the generation of almost unlimited amounts of supervised data and enable
researchers to isolate the performance of their algorithms on individual
simulated phenomena. Work on such data has shown that neural network based
models hold promise for modelling reading comprehension, something that we
will build upon here.  Historically, however, many similar approaches in
Computational Linguistics have failed to manage the transition from synthetic
data to real environments, as such closed worlds inevitably fail to
capture the complexity, richness, and noise of natural language
\cite{Winograd:1972:UNL}.

In this work we seek to directly address the lack of real natural language
training data by introducing a novel approach to building a supervised reading
comprehension data set. We observe that summary and paraphrase sentences, with
their associated documents, can be readily converted to context--query--answer
triples using simple entity detection and anonymisation algorithms.
Using this approach we have collected two new corpora of roughly a million news
stories with associated queries from the CNN and Daily Mail websites.

We demonstrate the efficacy of our new corpora by building novel deep learning
models for reading comprehension. These models draw on recent developments
for incorporating attention mechanisms into recurrent neural network architectures
\cite{Bahdanau:2014:NMT,Mnih:2014:RMVA,Gregor:2015:DRAW,Sukhbaatar:2015}. This allows a model to
focus on the aspects of a document that it believes will help it answer a
question, and also allows us to visualises its inference process.
We compare these neural models to a range of baselines and heuristic benchmarks
based upon a traditional frame semantic analysis provided by a state-of-the-art
natural language processing (NLP) pipeline. Our results indicate that the neural
models achieve a higher accuracy, and do so without any specific encoding of the
document or query structure.

\section{Supervised training data for reading comprehension}
\label{data}

The reading comprehension task naturally lends itself to a formulation as a
supervised learning problem. Specifically we seek to estimate the conditional
probability $p(a|c,q)$, where $c$ is a context document, $q$ a query relating to
that document, and $a$ the answer to that query.
For a focused evaluation we wish to be able to exclude additional information,
such as world knowledge gained from co-occurrence statistics, in order to test a
model's core capability to detect and understand the linguistic relationships
between entities in the context document.

Such an approach requires a large training corpus of document--query--answer
triples and until now such corpora have been limited to hundreds of examples and
thus mostly of use only for testing \cite{Richardson:2013:MCT}. This limitation
has meant that most work in this area has taken the form of unsupervised
approaches which use templates or syntactic/semantic analysers to extract
relation tuples from the document to form a knowledge graph that can be queried.

Here we propose a methodology for creating real-world, large scale supervised
training data for learning reading comprehension models. Inspired by work in
summarisation \cite{Svore:2007:CNN,Woodsend:2010:CNN}, we create two machine
reading corpora by exploiting online newspaper articles and their matching
summaries. We have collected 93k articles from the CNN\footnote{\url{www.cnn.com}} and 220k articles from
the Daily Mail\footnote{\url{www.dailymail.co.uk}} websites. Both news providers supplement their articles with a
number of bullet points, summarising aspects of the information contained in the
article. Of key importance is that these summary points are abstractive and do
not simply copy sentences from the documents.
We construct a corpus of document--query--answer triples by turning these
bullet points into Cloze \cite{Taylor:1953:CLOZE} style questions by replacing
one entity at a time with a placeholder. This results in a combined corpus of
roughly 1M data points (Table \ref{tab:corpora}).
Code to replicate our datasets---and to apply this method to other sources---is
available online\footnote{\url{http://www.github.com/deepmind/rc-data/}}.

\begin{table}[t]
\footnotesize
{\centering
  \begin{minipage}[t]{0.65\textwidth}
  \centering
  \begin{tabular}[t]{@{}l@{~}r@{~~}r@{~~}r@{}l@{}r@{~~}r@{~~}r@{}}
    \toprule
    & \multicolumn{3}{c}{{\bf CNN}} &\phantom{aa}& \multicolumn{3}{c}{{\bf
Daily Mail}} \\
    \cmidrule{2-4} \cmidrule{6-8}
    & train & valid & test && train & valid & test \\
    \midrule
    \# months    & 95       & 1     & 1     &&      56 & 1      & 1 \\
    \# documents &  90,266  & 1,220 & 1,093 && 196,961 & 12,148 & 10,397 \\
    \# queries   & 380,298  & 3,924 & 3,198 && 879,450 & 64,835 & 53,182 \\
    Max \# entities & 527   & 187   & 396   && 371     & 232    & 245 \\
    Avg \# entities & 26.4  & 26.5  & 24.5  && 26.5    & 25.5   & 26.0 \\
    Avg \# tokens  & 762    & 763   & 716   && 813     & 774    & 780  \\
    Vocab size & \multicolumn{3}{c}{{118,497}} && \multicolumn{3}{c}{{208,045}} \\
    \bottomrule
  \end{tabular}
  \captionof{table}{Corpus statistics. Articles were collected starting in
    April 2007 for CNN and June 2010 for the Daily Mail,
    both until the end of April~2015. Validation data is from March, test data from April~2015.
    Articles of over 2000 tokens and  queries whose answer entity did not appear in the context were filtered out.}
  \label{tab:corpora}
\end{minipage}
\hfill
\begin{minipage}[t]{0.33\textwidth}
\footnotesize
  \centering
  \begin{tabular}[t]{@{}l@{~~}r@{~~}r@{}}
    \toprule
    \textbf{Top N} & \multicolumn{2}{c}{{\bf Cumulative \%}} \\
    \cmidrule{2-3}
    & \textbf{CNN} & \textbf{Daily Mail} \\
    \midrule
    1  & 30.5 & 25.6 \\
    2  & 47.7 & 42.4 \\
    3  & 58.1 & 53.7 \\
    5  & 70.6 & 68.1 \\
    10 & 85.1 & 85.5 \\
    \bottomrule
  \end{tabular}
  \captionof{table}{Percentage of time that the correct answer is contained in the
  top $N$ most frequent entities in a given document.}
  \label{tab:cloze_hardness}
\end{minipage}
}
\end{table}

\subsection{Entity replacement and permutation}

\begin{table}[t]
  \footnotesize

  \begin{tabularx}{\textwidth}{@{}lY@{}l@{}Y@{}}
    \toprule

    & \textbf{Original Version} & \phantom{b}
                & \textbf{Anonymised Version} \\
    \midrule
    \multicolumn{4}{@{}l}{\textbf{Context}} \\
    & The BBC producer allegedly struck by Jeremy Clarkson will not
    press charges against the ``Top Gear'' host, his lawyer said Friday.
    Clarkson, who hosted one of the most-watched television shows in the world,
    was dropped by the BBC Wednesday after an internal investigation by the
    British broadcaster found he had subjected producer Oisin Tymon ``to an
    unprovoked physical and verbal attack.'' \dots
    &
    & the \textit{ent381} producer allegedly struck by \textit{ent212} will not
    press charges against the `` \textit{ent153} '' host , his lawyer said
    friday .  \textit{ent212} , who hosted one of the most - watched television
    shows in the world , was dropped by the \textit{ent381} wednesday after an
    internal investigation by the \textit{ent180} broadcaster found he had
    subjected producer \textit{ent193} `` to an unprovoked physical and verbal
    attack . '' \dots
    \\
    \midrule
    \multicolumn{4}{@{}l}{\textbf{Query}} \\
    & Producer \textbf{X} will not press charges against Jeremy Clarkson, his
    lawyer says.
    &
    & producer \textbf{X} will not press charges against \textit{ent212} , his
    lawyer says .
    \\
    \midrule
    \multicolumn{4}{@{}l}{\textbf{Answer}} \\
    & Oisin Tymon
    &
    & \textit{ent193} \\
    \bottomrule
  \end{tabularx}
  \caption{Original and anonymised version of a data point from the Daily Mail
  validation set. The anonymised entity markers are constantly permuted during
training and testing.}
  \label{tab:saccharin}
\end{table}

Note that the focus of this paper is to provide a corpus for evaluating a model's
ability to read and comprehend a single document, not world knowledge or
co-occurrence. To understand that distinction consider for instance the
following Cloze form queries (created from headlines in the Daily Mail
validation set):
\begin{inparaenum}[\itshape a\upshape)]
  \item The hi-tech bra that helps you beat breast \textbf{X};
  \item Could Saccharin help beat \textbf{X} ?;
  \item Can fish oils help fight prostate \textbf{X} ?
\end{inparaenum}
An ngram language model trained on the Daily Mail would easily correctly predict
that (\textbf{X} = \textit{cancer}), regardless of the contents of the context
document, simply because this is a very frequently cured entity in the Daily Mail
corpus.

To prevent such degenerate solutions and create a focused task we anonymise and
randomise our corpora with the following procedure,
\begin{inparaenum}[\itshape a\upshape)]
  \item use a coreference system to establish coreferents in each
    data point;
  \item replace all entities with abstract entity markers according to
    coreference;
  \item randomly permute these entity markers whenever a data point is loaded.
\end{inparaenum}

Compare the original and anonymised version of the example in Table
\ref{tab:saccharin}. Clearly a human reader can answer both queries correctly.
However in the anonymised setup the context document is required for answering
the query, whereas the original version could also be answered by someone with
the requisite background knowledge.
Therefore, following this procedure, the only remaining strategy for answering
questions is to do so by exploiting the context presented with each question.
Thus performance on our two corpora truly measures reading comprehension
capability.  Naturally a production system would benefit from using all
available information sources, such as clues through language and co-occurrence
statistics.

Table \ref{tab:cloze_hardness} gives an indication of the difficulty of the
task, showing how frequent the correct answer is contained in the top $N$ entity
markers in a given document. Note that our models don't distinguish between
entity markers and regular words. This makes the task harder and the models more
general.

\section{Models}
\label{models}

So far we have motivated the need for better datasets and tasks to evaluate the
capabilities of machine reading models. We proceed by describing a number of
baselines, benchmarks and new models to evaluate against this paradigm.  We
define two simple baselines, the majority baseline ({\tt maximum frequency})
picks the entity most frequently observed in the context document, whereas the
exclusive majority ({\tt exclusive frequency}) chooses the entity most
frequently observed in the context but not observed in the query. The idea
behind this exclusion is that the placeholder is unlikely to be mentioned twice
in a single Cloze form query.

\subsection{Symbolic Matching Models}

Traditionally, a pipeline of NLP models has been used for attempting question
answering, that is models that make heavy use of linguistic annotation,
structured world knowledge and semantic parsing and similar NLP pipeline
outputs.
Building on these approaches, we define a number of NLP-centric models for our
machine reading task.

\paragraph{Frame-Semantic Parsing}

Frame-semantic parsing attempts to identify predicates and their arguments,
allowing models access to information about ``who did what to whom''. Naturally
this kind of annotation lends itself to being exploited for question answering.
We develop a benchmark that makes use of frame-semantic annotations
which we obtained by parsing our model with a state-of-the-art frame-semantic
parser \cite{Das:2013:SRL,Hermann:2014:SRL}. As the parser makes extensive use
of linguistic information we run these benchmarks on the unanonymised version of
our corpora. There is no significant advantage in this as the frame-semantic
approach used here does not possess the capability to generalise through a
language model beyond exploiting one during the parsing phase.
Thus, the key objective of evaluating machine comprehension abilities is
maintained. Extracting entity-predicate triples---denoted as
$(e_1,V,e_2)$---from both the query $q$ and context document $d$, we attempt to
resolve queries using a number of rules with an increasing recall/precision
trade-off as follows (Table \ref{tab:fsp}).

\begin{table}[h]\footnotesize
  \centering
  \begin{tabular}{@{}rllll@{}}
    \toprule
    & Strategy & Pattern $\in q$ & Pattern $\in d$ & Example (Cloze / Context) \\
    \midrule
    1 & Exact match & $(p,V,y)$ & $(\bm{x},V,y)$ & X loves Suse / \textbf{Kim} loves Suse \\
    2 & be.01.V match & $(p,\textit{be.01.V},y)$ & $(\bm{x},\textit{be.01.V},y)$ & X is president / \textbf{Mike} is president \\
    3 & Correct frame & $(p,V,y)$ & $(\bm{x},V,z)$ & X won Oscar / \textbf{Tom} won Academy Award \\
    4 & Permuted frame & $(p,V,y)$ & $(y,V,\bm{x})$ & X met Suse / Suse met \textbf{Tom} \\
    5 & Matching entity & $(p,V,y)$ & $(\bm{x},Z,y)$ & X likes candy / \textbf{Tom} loves candy \\
    6 & Back-off strategy & \multicolumn{3}{l}{\textit{Pick the most frequent entity from the context that doesn't appear in the query}} \\
    \bottomrule
  \end{tabular}
  \caption{Resolution strategies using PropBank triples. $\bm{x}$ denotes the
    entity proposed as answer, $V$ is a fully qualified PropBank frame (e.g.
    \textit{give.01.V}). Strategies are ordered by precedence and answers
    determined accordingly. This heuristic algorithm was iteratively
    tuned on the validation data set.
    \label{tab:fsp}
  }
\end{table}

For reasons of clarity, we pretend that all PropBank triples are of the form
$(e_1,V,e_2)$. In practice, we take the argument numberings of the parser into
account and only compare like with like, except in cases such as the permuted
frame rule, where ordering is relaxed. In the case of multiple possible answers
from a single rule, we randomly choose one.

\paragraph{Word Distance Benchmark}

We consider another baseline that relies on word distance measurements. Here, we
align the placeholder of the Cloze form question with each possible entity in
the context document and calculate a distance measure between the question and the
context around the aligned entity.
This score is calculated by summing the distances of every word in $q$
to their nearest aligned word in $d$, where alignment is defined by matching
words either directly or as aligned by the coreference system. We tune the
maximum penalty per word ($m=8$) on the validation data.

\subsection{Neural Network Models}
Neural networks have successfully been applied to a range of tasks in NLP.
This includes classification tasks such as sentiment analysis
\cite{Kalchbrenner:2014:DCNN} or POS tagging \cite{Collobert:2011:NLP}, as well
as generative problems such as language modelling or machine translation
\cite{Sutskever:2014:SSLNN}.
We propose three neural models for estimating the probability of word type $a$
from document $d$ answering query $q$:
\begin{align*}
  p(a | d, q) &\propto \exp \left(W(a) g(d,q) \right), \quad\text{s.t. } a \in
  V,
\end{align*}
where $V$ is the vocabulary\footnote{The vocabulary includes all the word types
  in the documents, questions, the entity maskers, and the question unknown
  entity marker.},
and $W(a)$ indexes row $a$ of weight matrix $W$ and through a slight abuse of
notation word types double as indexes. Note that we do not privilege entities or
variables, the model must learn to differentiate these in the input sequence.
The function $g(d,q)$ returns a vector embedding of a document and query pair.

\paragraph{The Deep LSTM Reader}
Long short-term memory (LSTM, \cite{Hochreiter:1997:LSTM}) networks have
recently seen considerable success in tasks such as machine translation and
language modelling \cite{Sutskever:2014:SSLNN}. When used for translation, Deep
LSTMs \cite{Graves:2012:SSLRNN} have shown a remarkable ability to embed long
sequences into a vector representation which contains enough information to
generate a full translation in another language. Our first neural model for
reading comprehension tests the ability of Deep LSTM encoders to handle
significantly longer sequences. We feed our documents one word at a time into
a Deep LSTM encoder, after a delimiter we then also feed the query into the
encoder. Alternatively we also experiment with processing the query then the
document. The result is that this model processes each document query pair as a
single long sequence. Given the embedded document and query the network
predicts which token in the document answers the query.

We employ a Deep LSTM cell with skip connections from each input $x(t)$ to every
hidden layer, and from every hidden layer to the output $y(t)$:
\begin{align*}
  x'(t,k) &= x(t)||y'(t,k-1), \quad\quad y(t) = y'(t,1) || \ldots || y'(t,K) \\
  \igate(t,k) &= \sigma\left(\wtmat{kx}{\igate} x'(t,k) + \wtmat{kh}{\igate} h(t-1,k) + \wtmat{k\state}{\igate} \state(t-1,k)  + b_{k\igate}\right)\\
  \fgate(t,k) &= \sigma\left(\wtmat{kx}{\fgate} x(t) + \wtmat{kh}{\fgate} h(t-1,k) + \wtmat{k\state}{\fgate} \state(t-1,k) + b_{k\fgate} \right)\\
  \state(t,k) &= \fgate(t,k) \state(t-1,k) + \igate(t,k) \tanh \left(\wtmat{kx}{\state} x'(t,k) + \wtmat{kh}{\state} h(t-1,k) + b_{k\state} \right) \\
  \ogate(t,k) &= \sigma\left(\wtmat{kx}{\ogate} x'(t,k) + \wtmat{kh}{\ogate} h(t-1,k) + \wtmat{k\state}{\ogate} \state(t,k) + b_{k\ogate}\right)\\
  h(t,k) &= \ogate(t,k) \tanh\left(\state(t,k)\right)\\
  y'(t,k) &= \wtmat{k}{y}h(t,k) + b_{ky}
\end{align*}
where $||$ indicates vector concatenation $h(t,k)$ is the hidden state for layer
$k$ at time $t$, and $\igate$, $\fgate$, $\ogate$ are the input, forget, and
output gates respectively.
Thus our Deep LSTM Reader is defined by $g^{\text{\tiny LSTM}}(d,q) = y(|d|+|q|)$ with input $x(t)$ the concatenation of $d$ and $q$ separated by the delimiter $|||$.

\paragraph{The Attentive Reader}

\newcommand{\attnIn}{m}
\newcommand{\attnU}{u}
\newcommand{\attnOver}{y}
\newcommand{\attnMix}{r}
\newcommand{\attnMid}{z}
\newcommand{\attnScore}{s}
\newcommand{\fwd}[1]{\overrightarrow{#1}}
\newcommand{\back}[1]{\overleftarrow{#1}}
\newcommand{\softmax}[2]{\frac{\exp\left(#1\right)}{#2}}

\begin{figure}
\centering
  \begin{subfigure}[b]{0.49\textwidth}
    \centering
    \includegraphics[scale=0.44]{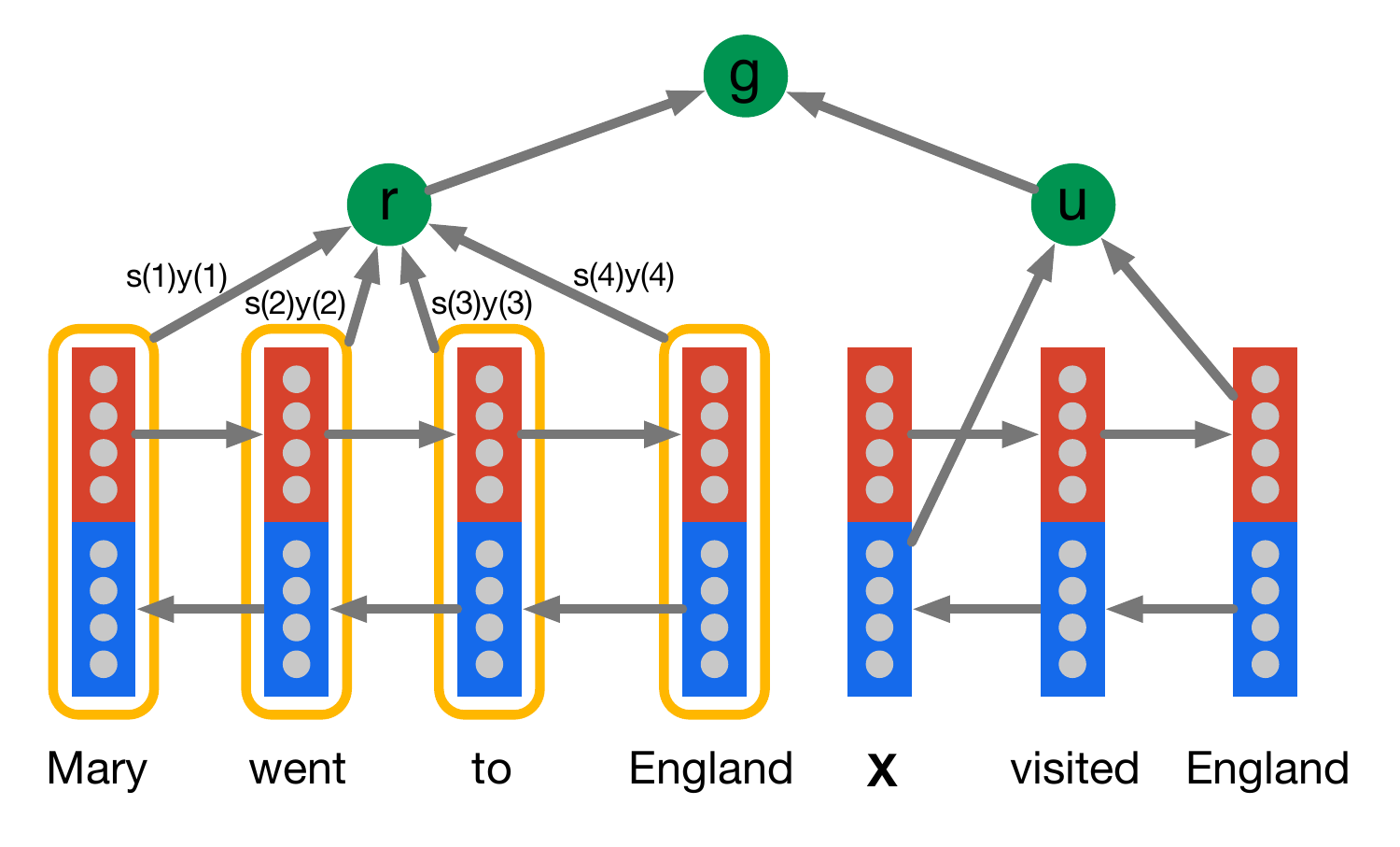}
    \caption{Attentive Reader.}
  \end{subfigure}
  \begin{subfigure}[b]{0.49\textwidth}
    \centering
    \includegraphics[scale=0.44]{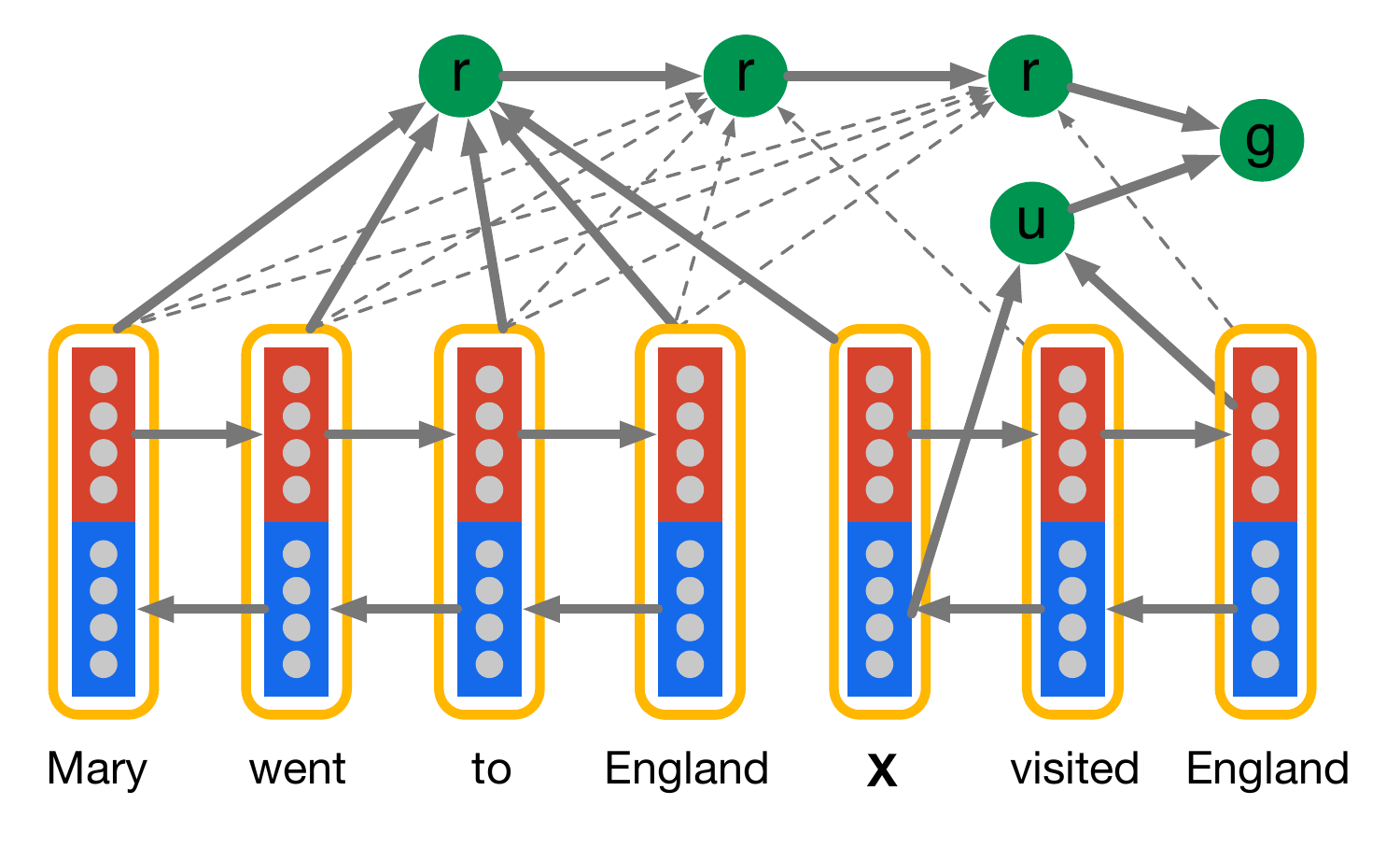}
    \caption{Impatient Reader.}
  \end{subfigure}
  \begin{subfigure}[b]{1.0\textwidth}
    \centering
    \includegraphics[scale=0.44]{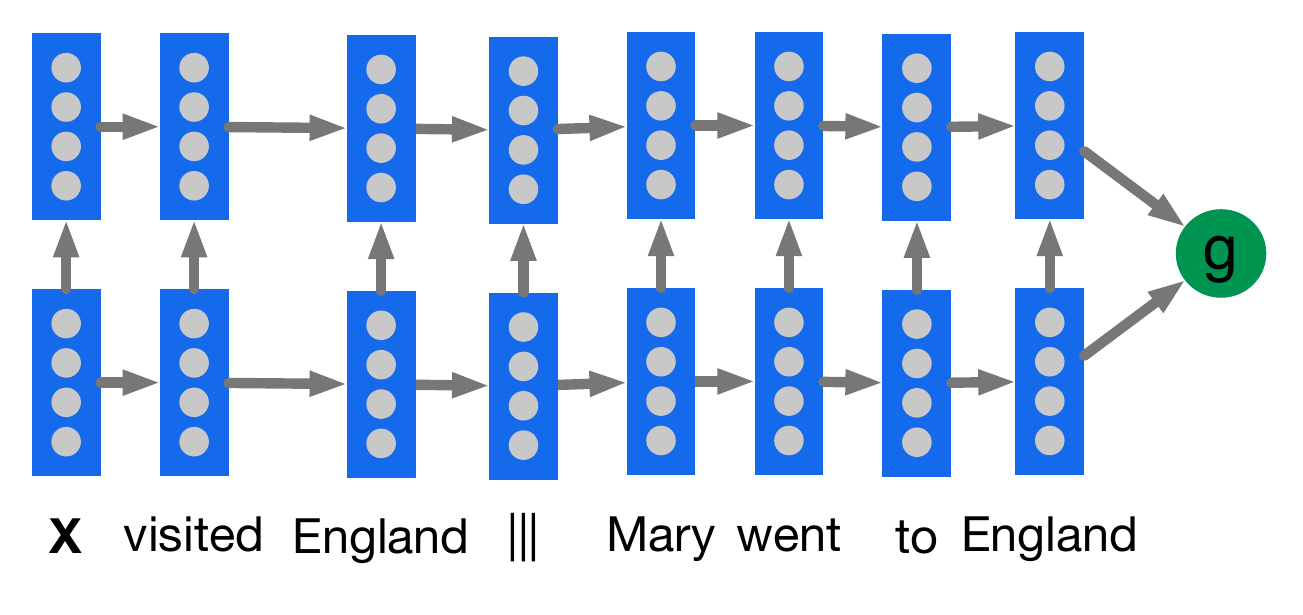}
    \caption{A two layer Deep LSTM Reader with the question encoded before
             the document.}
  \end{subfigure}
  \caption{Document and query embedding models.}
\label{fig:models}
\end{figure}

The Deep LSTM Reader must propagate dependencies over long distances in order to
connect queries to their answers. The fixed width hidden vector forms a
bottleneck for this information flow that we propose to circumvent using an
attention mechanism inspired by recent results in translation and image
recognition \cite{Bahdanau:2014:NMT,Mnih:2014:RMVA}.
This attention model first encodes the document and the query using separate
bidirectional single layer LSTMs \cite{Graves:2012:SSLRNN}.

We denote the
outputs of the forward and backward LSTMs as $\fwd{y}(t)$ and $\back{y}(t)$
respectively.  The encoding $u$ of a query of length $|q|$ is formed by the
concatenation of the final forward and backward outputs,
$u = \fwd{y_q}(|q|)\,\, ||\,\, \back{y_q}(1).$

For the document the composite output for each token at position $t$ is,
$y_d(t) = \fwd{y_d}(t)\,\, ||\,\, \back{y_d}(t).$
The representation $r$ of the document $d$ is formed by a weighted sum of these
output vectors. These weights are interpreted as the degree to which the network
attends to a particular token in the document when answering the query:
\begin{align*}
  \attnIn(t)    &= \tanh\left(\wtmat{\attnOver}{\attnIn} \attnOver_d(t)
                   + \wtmat{\attnU}{\attnIn} \attnU\right),\\
  \attnScore(t) &\propto \exp \left(\mathrm{w}_{\attnIn\attnScore}^\intercal
                         \attnIn(t) \right),\\
  \attnMix   &= \attnOver_d \attnScore,
\end{align*}
where we are interpreting $y_d$ as a matrix with each column being the composite
representation $y_d(t)$ of document token $t$.
The variable $\attnScore(t)$ is the normalised attention at token $t$. Given
this attention score the embedding of the document $\attnMix$ is computed as the
weighted sum of the token embeddings.
The model is completed with the definition of the joint document and query
embedding via a non-linear combination:
\begin{align*}
  g^{\text{\tiny AR}}(d,q) = \tanh \left(\wtmat{\attnMix}{g} \attnMix
                             + \wtmat{\attnU}{g} \attnU \right).
\end{align*}

The Attentive Reader can be viewed as a generalisation of the application of
Memory Networks to question answering \cite{Weston:2014:MN}. That model employs
an attention mechanism at the sentence level where each sentence is represented
by a bag of embeddings. The Attentive Reader employs a finer grained token
level attention mechanism where the tokens are embedded given their entire
future and past context in the input document.

\paragraph{The Impatient Reader}
The Attentive Reader is able to focus on the passages of a context document
that are most likely to inform the answer to the query. We can go further by
equipping the model with the ability to reread from the document as each query
token is read. At each token $i$ of the query $q$ the model computes a document
representation vector $r(i)$ using the bidirectional embedding $y_q(i) =
\fwd{y_q}(i)\,\, ||\,\, \back{y_q}(i)$:
\begin{align*}
  \attnIn(i, t)   &= \tanh\left(\wtmat{d}{\attnIn} \attnOver_d(t)
                   + \wtmat{\attnMix}{\attnIn} \attnMix(i-1)
                   + \wtmat{q}{\attnIn} y_q(i) \right)
                , \quad 1 \leq i \leq |q|,  \\
  \attnScore(i,t) &\propto \exp \left(\mathrm{w}_{\attnIn\attnScore}^\intercal
                   \attnIn(i,t) \right),\\
  \attnMix(0)     &= \mathbf{r_0}, \quad
                   \attnMix(i)     = \attnOver_d^\intercal \attnScore(i) +
                   \tanh\left(\wtmat{\attnMix}{\attnMix}\attnMix(i-1)\right)
  \quad 1 \leq i \leq |q|.
\end{align*}
The result is an attention mechanism that allows the model to recurrently
accumulate information from the document as it sees each query token, ultimately
outputting a final joint document query representation for the answer prediction,
\begin{align*}
  g^{\text{\tiny IR}}(d,q) = \tanh \left(\wtmat{\attnMix}{g} \attnMix(|q|) +
                             \wtmat{q}{g} \attnU \right).
\end{align*}

\section{Empirical Evaluation}
\label{experiments}

Having described a number of models in the previous section, we next evaluate
these models on our reading comprehension corpora. Our hypothesis is that neural models should in principle be well suited for this
task. However, we argued that simple recurrent models such as the LSTM
probably have insufficient expressive power for solving tasks that require
complex inference. We expect that the attention-based models would therefore
outperform the pure LSTM-based approaches.

Considering the second dimension of our investigation, the comparison of
traditional versus neural approaches to NLP, we do not have a strong prior
favouring one approach over the other. While numerous publications in the past
few years have demonstrated neural models outperforming classical methods, it
remains unclear how much of that is a side-effect of the language modelling
capabilities intrinsic to any neural model for NLP. The entity anonymisation and
permutation aspect of the task presented here may end up levelling the playing
field in that regard, favouring models capable of dealing with syntax rather
than just semantics.

With these considerations in mind, the experimental part of this paper is
designed with a three-fold aim. First, we want to establish the difficulty of our
machine reading task by applying a wide range of models to it. Second, we compare
the performance of parse-based methods versus that of neural models. Third,
within the group of neural models examined, we want to determine what each
component contributes to the end performance; that is, we want to analyse the
extent to which an LSTM can solve this task, and to what extent various
attention mechanisms impact performance.

\newcommand{\ee}[1][]{\text{\sc{e}#1}}

\begin{figure}[t]
{\centering
\begin{minipage}{0.48\textwidth}
  \footnotesize
  \centering
  \begin{tabular}{@{}l@{}rr@{}l@{}rr@{}}
    \toprule
    & \multicolumn{2}{c}{CNN} &\phantom{aa}& \multicolumn{2}{c}{Daily Mail} \\
    \cmidrule{2-3} \cmidrule{5-6}
    & valid & test && valid & test \\
    \midrule
    Maximum frequency              & 30.5 & 33.2 && 25.6 & 25.5 \\
    Exclusive frequency            & 36.6 & 39.3 && 32.7 & 32.8 \\
    Frame-semantic model~~         & 36.3 & 40.2 && 35.5 & 35.5 \\
    Word distance model            & 50.5 & 50.9 && 56.4 & 55.5 \\
    \midrule
    Deep LSTM Reader               & 55.0 & 57.0 && 63.3 & 62.2 \\
    Uniform Reader                 & 39.0 & 39.4 && 34.6 & 34.4 \\
    Attentive Reader               & 61.6 & 63.0 && {\bf 70.5} & {\bf 69.0} \\
    Impatient Reader               & {\bf 61.8} & {\bf 63.8} && 69.0 & 68.0 \\
    \bottomrule
  \end{tabular}
  \captionof{table}{Accuracy of all the models and benchmarks on the CNN and
                    Daily Mail datasets.
                    The Uniform Reader baseline sets all of the $m(t)$
                    parameters to be equal.}
  \label{tab:main_results}
\end{minipage}
\hspace{0.15in}
\begin{minipage}{0.47\textwidth}
\centering
\includegraphics[scale=0.24,trim=0 0.3cm 0 1.3cm,clip=true]{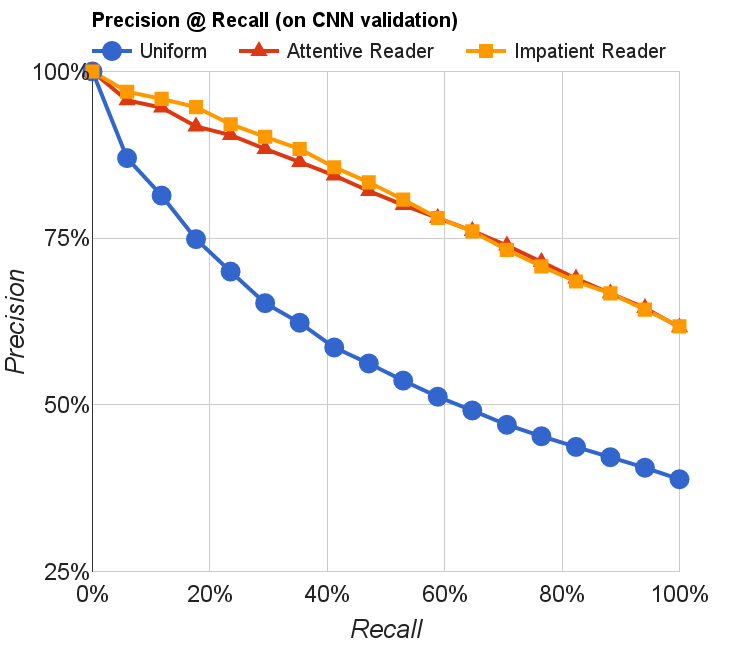}\\[-0.3em]
\captionof{figure}{Precision@Recall for the attention models on the CNN validation data.}
\label{fig:patr}
\end{minipage}
}
\end{figure}

 All model hyperparameters were tuned on the respective validation sets of the
 two corpora.\footnote{For the Deep LSTM Reader, we consider hidden layer sizes
   ${[64,128,\underline{256}]}$, depths ${[1,\underline{2},4]}$, initial
   learning rates ${[1\ee{-}3,5\ee{-}4,\underline{1\ee{-}4},5\ee{-}5]}$, batch
   sizes ${[16,\underline{32}]}$ and dropout $[0.0,\underline{0.1},0.2]$.  We
   evaluate two types of feeds. In the \textit{cqa} setup we feed first the
   context document and subsequently the question into the encoder, while the
   \textit{qca} model starts by feeding in the question followed by the context
   document. We report results on the best model (underlined hyperparameters,
   \textit{qca} setup).  For the attention models we consider hidden layer sizes
   $[64,128,256]$, single layer,  initial learning rates
 $[1\ee{-}4,5\ee{-}5,2.5\ee{-}5,1\ee{-}5]$, batch sizes $[8,16,32]$ and dropout
 $[0,0.1,0.2,0.5]$. For all models we used asynchronous RmsProp
 \cite{Tieleman:2012:RMSPROP} with a momentum of $0.9$ and a decay of $0.95$.
 See Appendix \ref{app:hyper} for more details of the experimental setup.}
Our experimental results are in Table \ref{tab:main_results}, with the
Attentive and Impatient Readers performing best across both datasets.

\paragraph{Frame-semantic benchmark}

While the one frame-semantic model proposed in this paper is clearly a
simplification of what could be achieved with annotations from an NLP pipeline,
it does highlight the difficulty of the task when approached from a symbolic NLP
perspective.

Two issues stand out when analysing the results in detail. First, the
frame-semantic pipeline has a poor degree of coverage with many relations not
being picked up by our PropBank parser as they do not adhere to the default
predicate-argument structure. This effect is exacerbated by the type of language
used in the highlights that form the basis of our datasets.
The second issue is that the frame-semantic approach does not trivially scale to
situations where several sentences, and thus frames, are required to answer a
query. This was true for the majority of queries in the dataset.

\paragraph{Word distance benchmark}

More surprising perhaps is the relatively strong performance of the word
distance benchmark, particularly relative to the frame-semantic benchmark, which
we had expected to perform better. Here, again, the nature of the datasets used
can explain aspects of this result. Where the frame-semantic model suffered due
to the language used in the highlights, the word distance model benefited.
Particularly in the case of the Daily Mail dataset, highlights frequently have
significant lexical overlap with passages in the accompanying article,
which makes it easy for the word distance benchmark.
For instance the query ``\textit{Tom Hanks is friends with {\bf X}'s
manager, Scooter Brown}'' has the phrase ``\textit{... turns out he is good
friends with Scooter Brown, manager for Carly Rae Jepson}'' in the context. The
word distance benchmark correctly aligns these two while the frame-semantic
approach fails to pickup the friendship or management relations when parsing
the query.
We expect that on other types of machine reading data where questions rather
than Cloze queries are used this particular model would perform significantly
worse.

\paragraph{Neural models}

Within the group of neural models explored here, the results paint a clear
picture with the Impatient and the Attentive Readers outperforming all other
models. This is consistent with our hypothesis that attention is a key
ingredient for machine reading and question answering due to the need to
propagate information over long distances.  The Deep LSTM Reader
performs surprisingly well, once again demonstrating that this simple sequential
architecture can do a reasonable job of learning to abstract long sequences,
even when they are up to two thousand tokens in length.  However this model does
fail to match the performance of the attention based models, even though these
only use single layer LSTMs.\footnote{Memory constraints prevented us from
experimenting with deeper Attentive Readers.}

The poor results of the Uniform Reader support our hypothesis of
the significance of the attention mechanism in the Attentive model's
performance as the only difference between these models is that the attention
variables are ignored in the Uniform Reader. The precision@recall statistics in
Figure~\ref{fig:patr} again highlight the strength of the attentive approach.

\begin{figure}[t]
  \centering
  \includegraphics[scale=0.30]{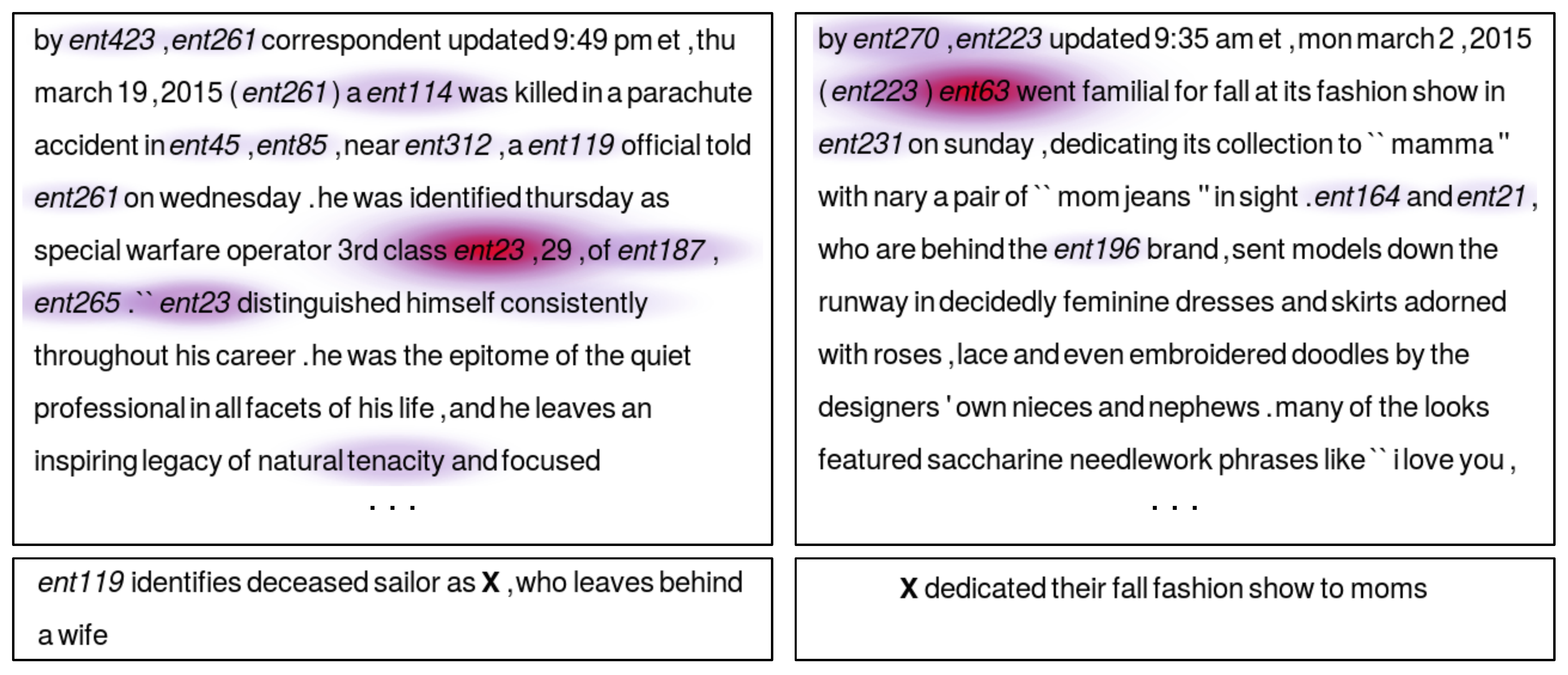}
  \caption{Attention heat maps from the Attentive Reader for two
    correctly answered validation set queries (the correct answers are
    \textit{ent23} and \textit{ent63}, respectively). Both examples
           require significant lexical generalisation and co-reference
           resolution in order to be answered correctly by a given model.}
  \label{fig:heatmaps}
\end{figure}

We can visualise the attention mechanism as a heatmap over a context document to
gain further insight into the models' performance. The highlighted words show
which tokens in the document were attended to by the model. In addition we must
also take into account that the vectors at each token integrate long range
contextual information via the bidirectional LSTM encoders.
Figure \ref{fig:heatmaps} depicts heat maps for two queries that were correctly
answered by the Attentive Reader.\footnote{Note that these examples were chosen
as they were short, the average CNN validation document contained 763 tokens and
27 entities, thus most instances were significantly harder to answer than these
examples.} In both cases confidently arriving at the correct answer requires the
model to perform both significant lexical generalsiation, e.g.\ `killed'
$\rightarrow$ `deceased', and co-reference or anaphora resolution, e.g.\
`{\em ent119} was killed' $\rightarrow$ `he was identified.' However it is also
clear that the model is able to integrate these signals with rough heuristic
indicators such as the proximity of query words to the candidate answer.

\section{Conclusion}
\label{conclusions}
The supervised paradigm for training machine reading and comprehension models
provides a promising avenue for making progress on the path to building full
natural language understanding systems. We have demonstrated a methodology for
obtaining a large number of document-query-answer triples and shown that
recurrent and attention based neural networks provide an effective modelling
framework for this task.
Our analysis indicates that the Attentive and Impatient Readers are able to
propagate and integrate semantic information over long distances. In particular
we believe that the incorporation of an attention mechanism is the key
contributor to these results.

The attention mechanism
that we have employed is just one instantiation of a very general idea which
can be further exploited. However, the incorporation of world knowledge and
multi-document queries will also require the development of attention and
embedding mechanisms whose complexity to query does not scale linearly with the
data set size.
There are still many queries requiring complex inference and long range reference resolution that our models are not yet able to answer. As such our data provides a scalable challenge that should support NLP research into the future. Further, significantly bigger training data sets can be acquired using the techniques we have described, undoubtedly allowing us to train more expressive and accurate models.

\newpage
\bibliographystyle{unsrt}
\bibliography{gilaaBibliography}

\newpage
\appendix
\section{Model hyperparameters}\label{app:hyper}

The precise hyperparameters used for the various attentive models are as in
Table \ref{tab:hyper}. All models were trained using asynchronous RmsProp
\cite{Tieleman:2012:RMSPROP} with a momentum of $0.9$ and a decay of $0.95$.

\begin{table}[h]
  \centering
  \begin{tabular}{@{}lrrrr@{}}
    \toprule
    Model & Hidden Size & Learning Rate & Batch Size & Dropout \\
    \midrule
    Uniform, CNN & 256 & 5\ee{-}5 & 32 & 0.2 \\
    Attentive, CNN & 256 & 5\ee{-}5 & 32 & 0.2 \\
    Impatient, CNN & 256 & 5\ee{-}5 & 32 & 0.3 \\
    \midrule
    Uniform, Daily Mail & 256 & 5\ee{-}5 & 32 & 0.2 \\
    Attentive, Daily Mail & 256 & 2.5\ee{-}5 & 32 & 0.1 \\
    Impatient, Daily Mail & 256 & 5\ee{-}5 & 32 & 0.1 \\
    \bottomrule
  \end{tabular}
  \caption{Model hyperparameters}
  \label{tab:hyper}
\end{table}

\section{Performance across document length}\label{app:length}

\begin{figure}[h]
{\centering
  \begin{minipage}[t]{0.49\textwidth}
  \centering
  \includegraphics[scale=0.27,trim=0 0.1cm 0 1.8cm,clip=true]{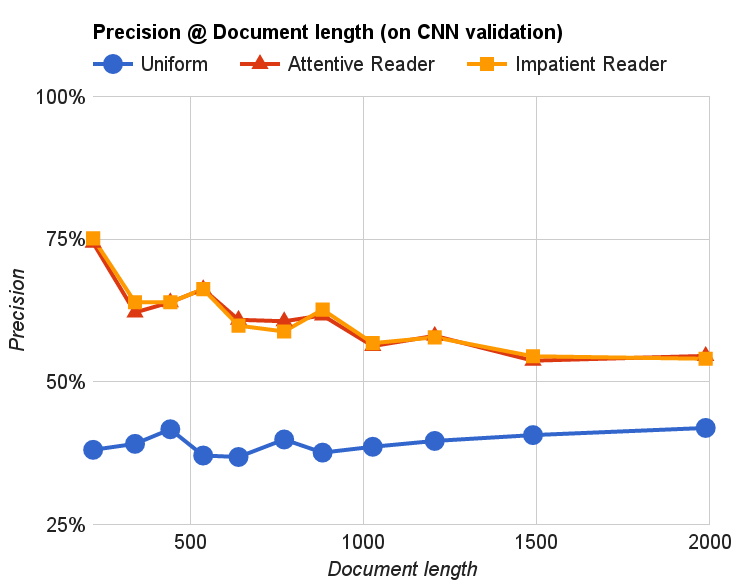}
  \captionof{figure}{Precision@Document Length for the attention models on the CNN
    validation data. The chart shows the precision for each decile in document lengths
  across the corpus as well as the precision for the 5\% longest articles.}
  \label{fig:patldec}
\end{minipage}
\hfill
\begin{minipage}[t]{0.49\textwidth}
\centering
  \includegraphics[scale=0.27,trim=0 0.1cm 0 1.8cm,clip=true]{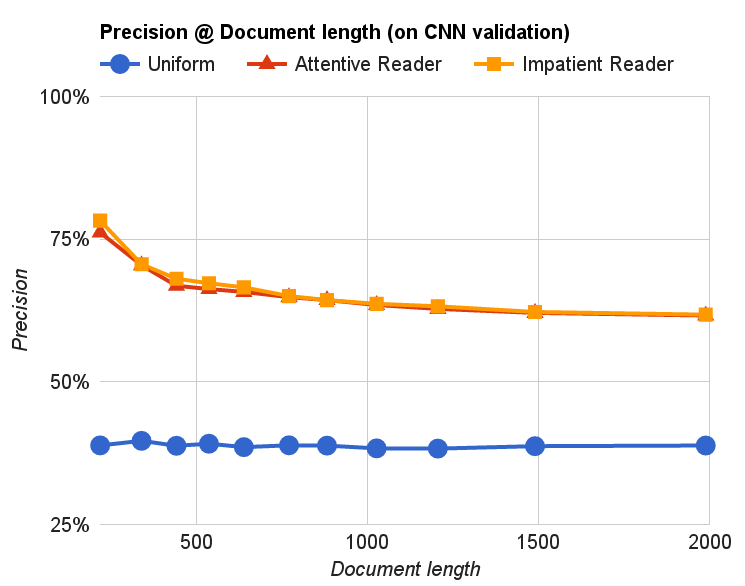}
  \captionof{figure}{Aggregated precision for documents up to a certain lengths.
  The points mark the $i^{th}$ decile in document lengths across the corpus.}
\label{fig:patl}
\end{minipage}
}
\end{figure}

To understand how the model performance depends on the size of the context, we
plot performance versus document lengths in Figures \ref{fig:patldec} and
\ref{fig:patl}. The first figure (Fig. \ref{fig:patldec}) plots a sliding window
of performance across document length, showing that performance of the attentive
models degrades slightly as documents increase in length. The second figure
(Fig. \ref{fig:patl}) shows the cumulative performance with documents up to
length $N$, showing that while the length does impact the models' performance,
that effect becomes negligible after reaching a length of \texttildelow 500~tokens.

\section{Additional Heatmap Analysis}

We expand on the analysis of the attention mechanism presented in the paper by
including visualisations for additional queries from the CNN validation dataset
below. We consider examples from the Attentive Reader as well as the Impatient
Reader in this appendix.

\subsection{Attentive Reader}

\begin{figure}[h]
  \centering
  \includegraphics[scale=0.30,clip=true,trim=3cm 17cm 3cm 2cm]{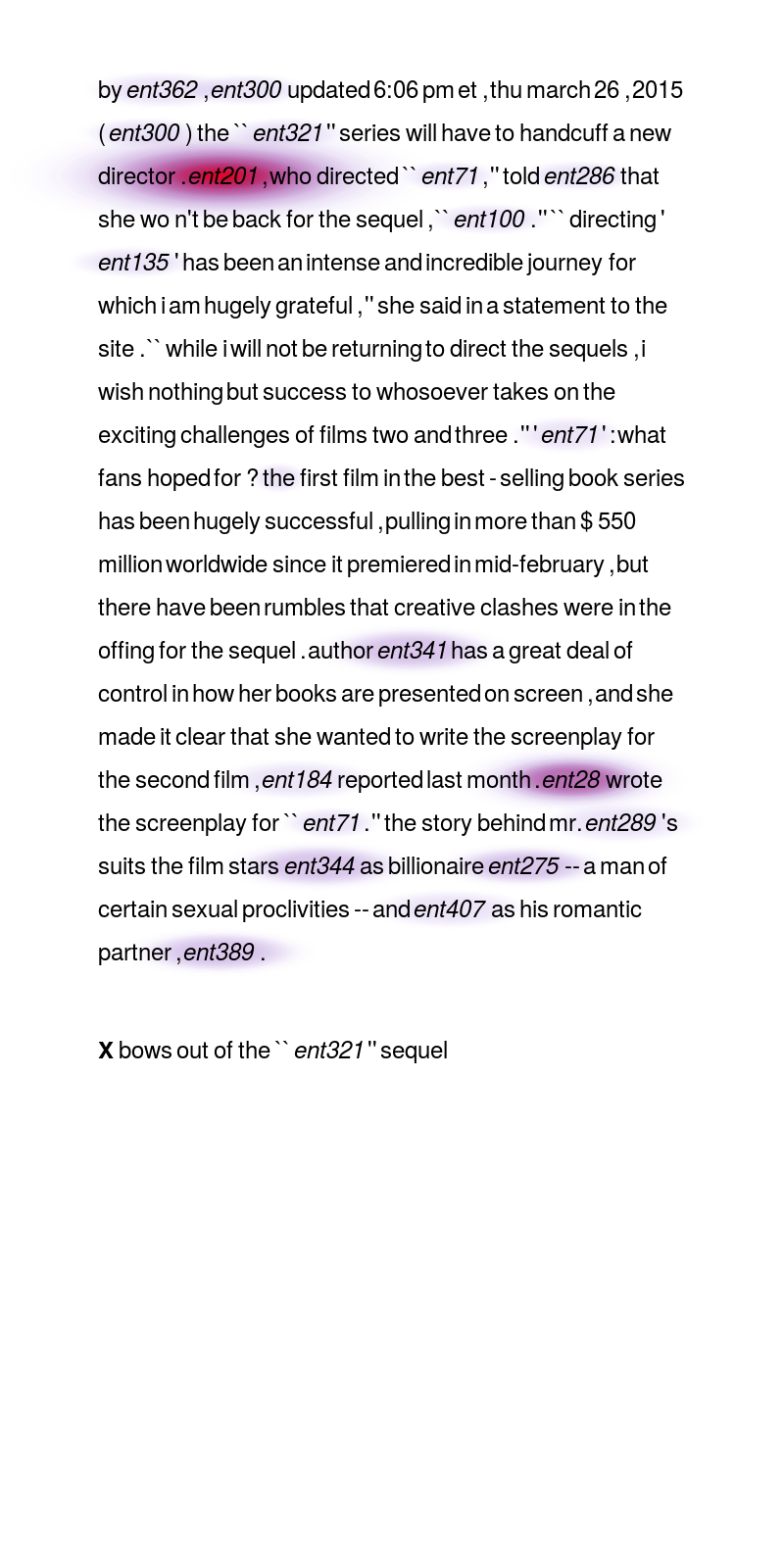}%
  \includegraphics[scale=0.30,clip=true,trim=2cm 17cm 2cm 2cm]{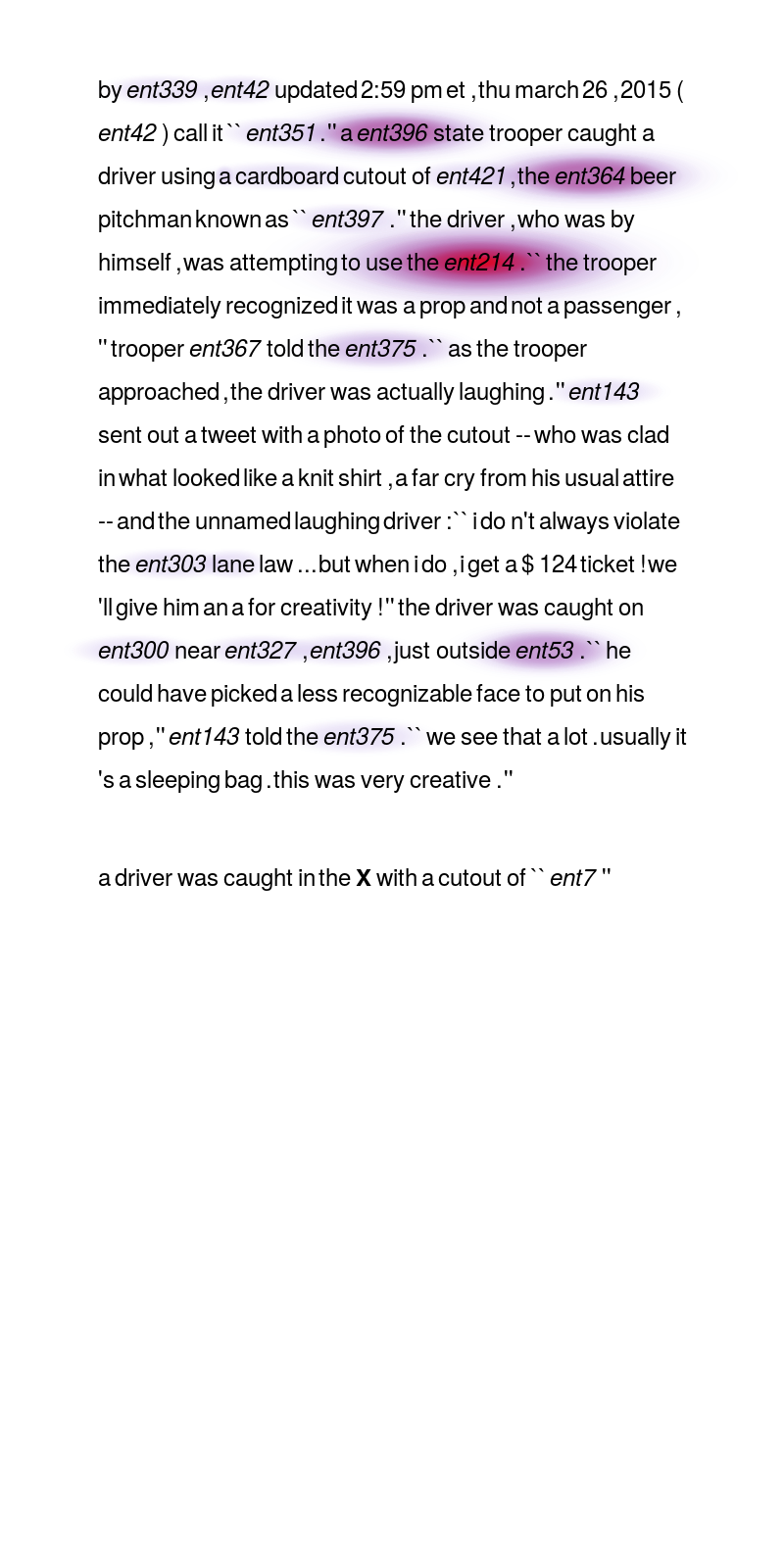}
  \caption{
	   Attention heat maps from the Attentive Reader for two more
           correctly answered validation set queries. Both examples
           require significant lexical generalisation and co-reference
           resolution to find the correct answers \textit{ent201} and
           \textit{ent214}, respectively.
       }
  \label{fig:heatmapsA}
\end{figure}

\begin{figure}[t]
  \centering
  \includegraphics[scale=0.30,clip=true,trim=3cm 27cm 2.2cm 2cm]{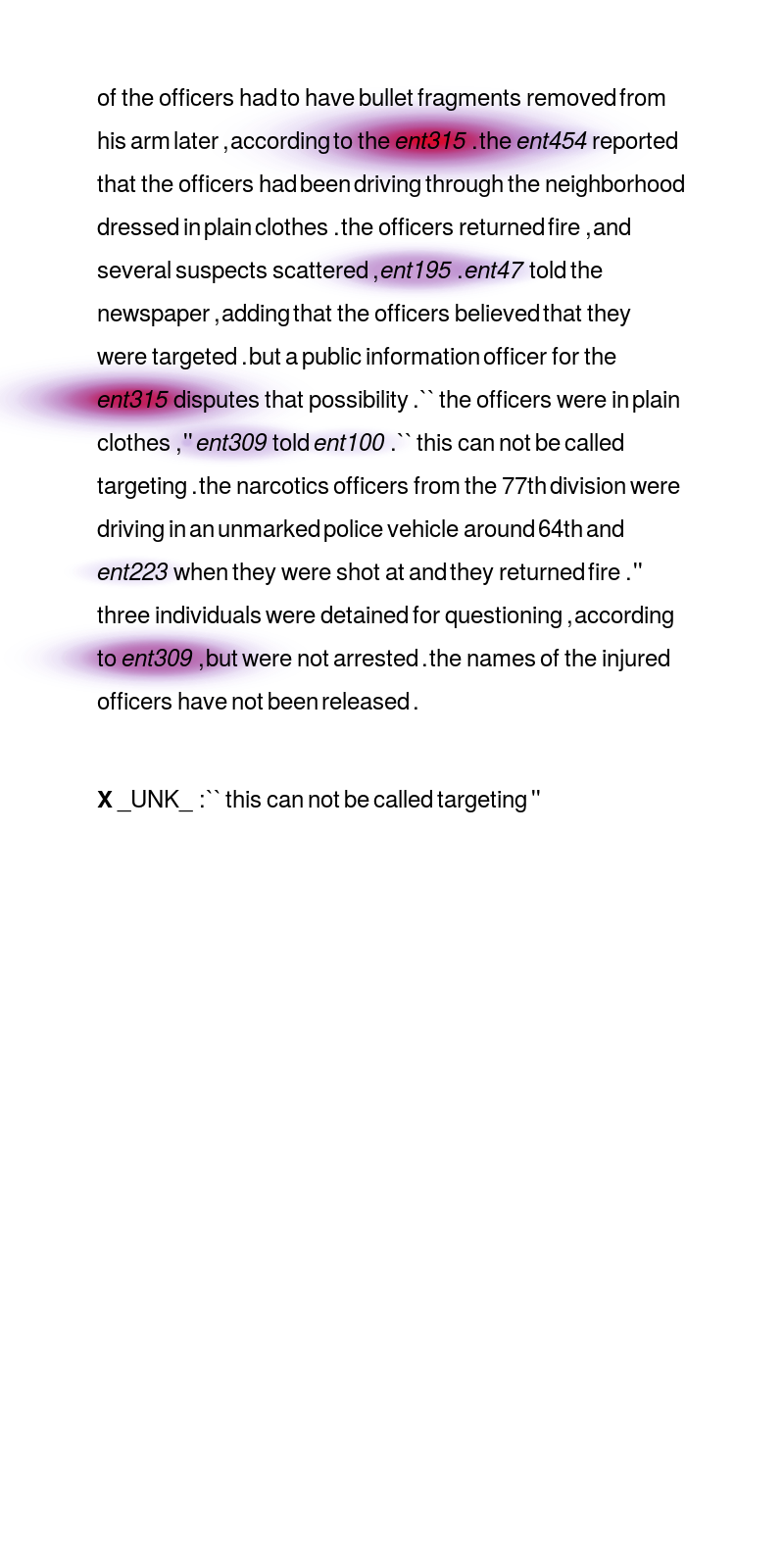}%
  \includegraphics[scale=0.30,clip=true,trim=2cm 27cm 3cm 2cm]{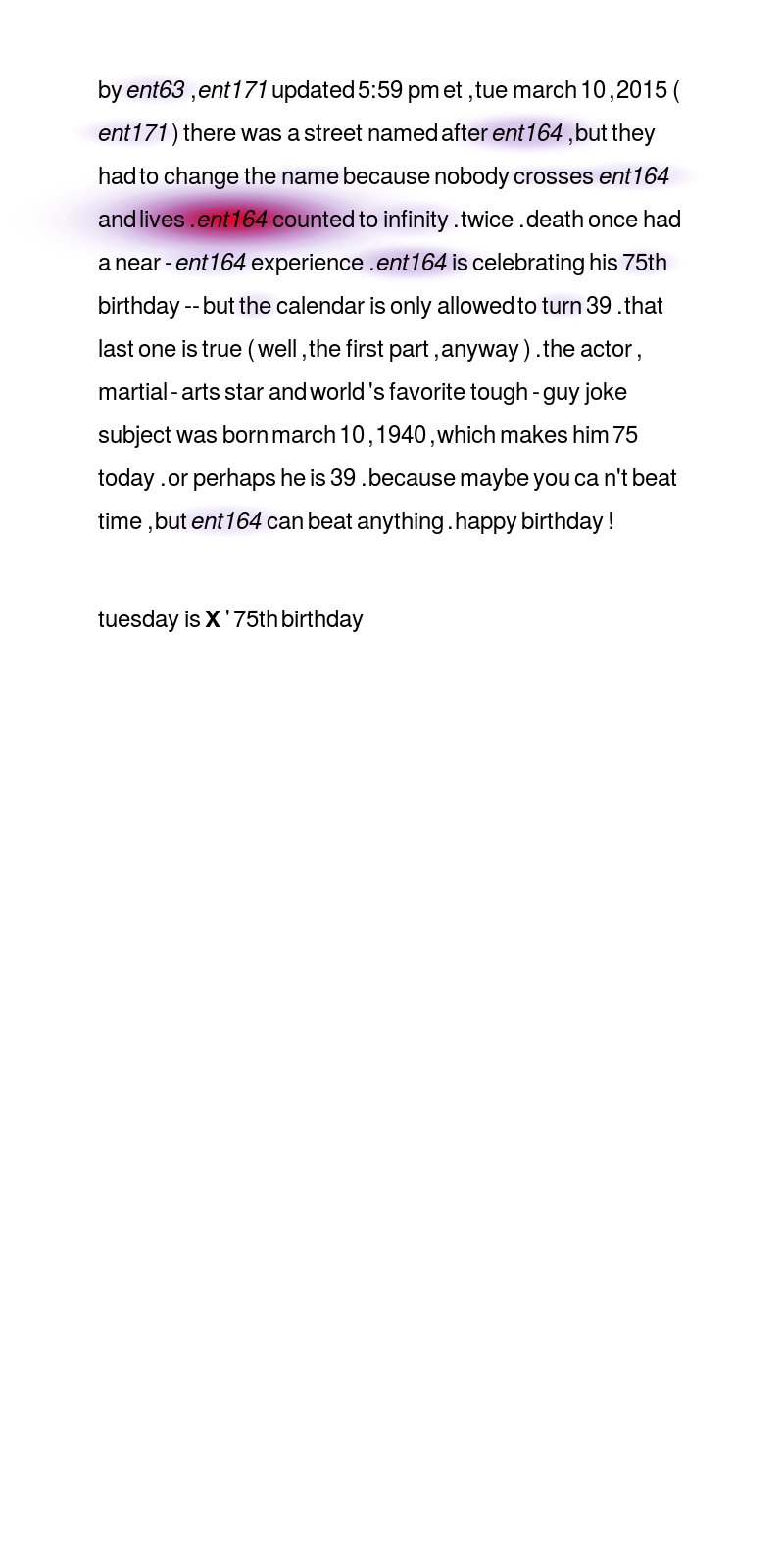}
  \caption{
    Two more
    correctly answered validation set queries. The left example (entity
    \textit{ent315}) requires correctly attributing the quote, which does not
    appear trivial with a number of other candidate entities in the vicinity.
    The right hand side shows our model is not afraid of Chuck Norris
    (\textit{ent164}).
  }
  \label{fig:heatmapsB}
\end{figure}

\paragraph{Positive Instances}
Figure \ref{fig:heatmapsA} shows two positive examples from the CNN validation
set that require reasonable levels of lexical generalisation and co-reference in
order to be answered.
The first query in Figure \ref{fig:heatmapsB} contains strong lexical cues
through the quote, but requires identifying the entity quoted, which is
non-trivial in the context document. The final positive example (also in Figure
\ref{fig:heatmapsB}) demonstrates the fearlessness of our model.

\begin{figure}[t]
  \centering
  \includegraphics[scale=0.3,clip=true,trim=3cm 32cm 2cm 2cm]{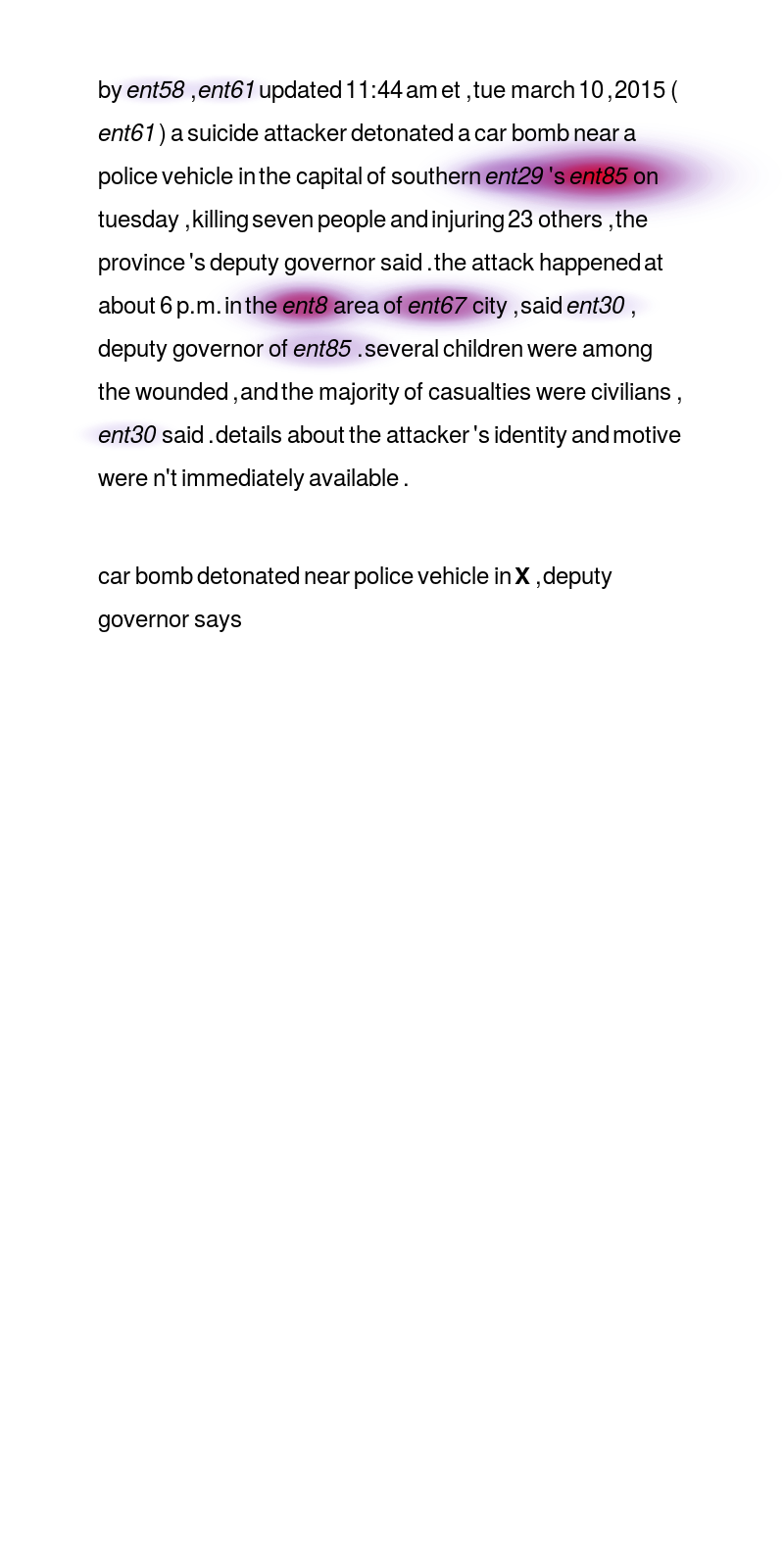}%
  \includegraphics[scale=0.3,clip=true,trim=2cm 32cm 3cm 2cm]{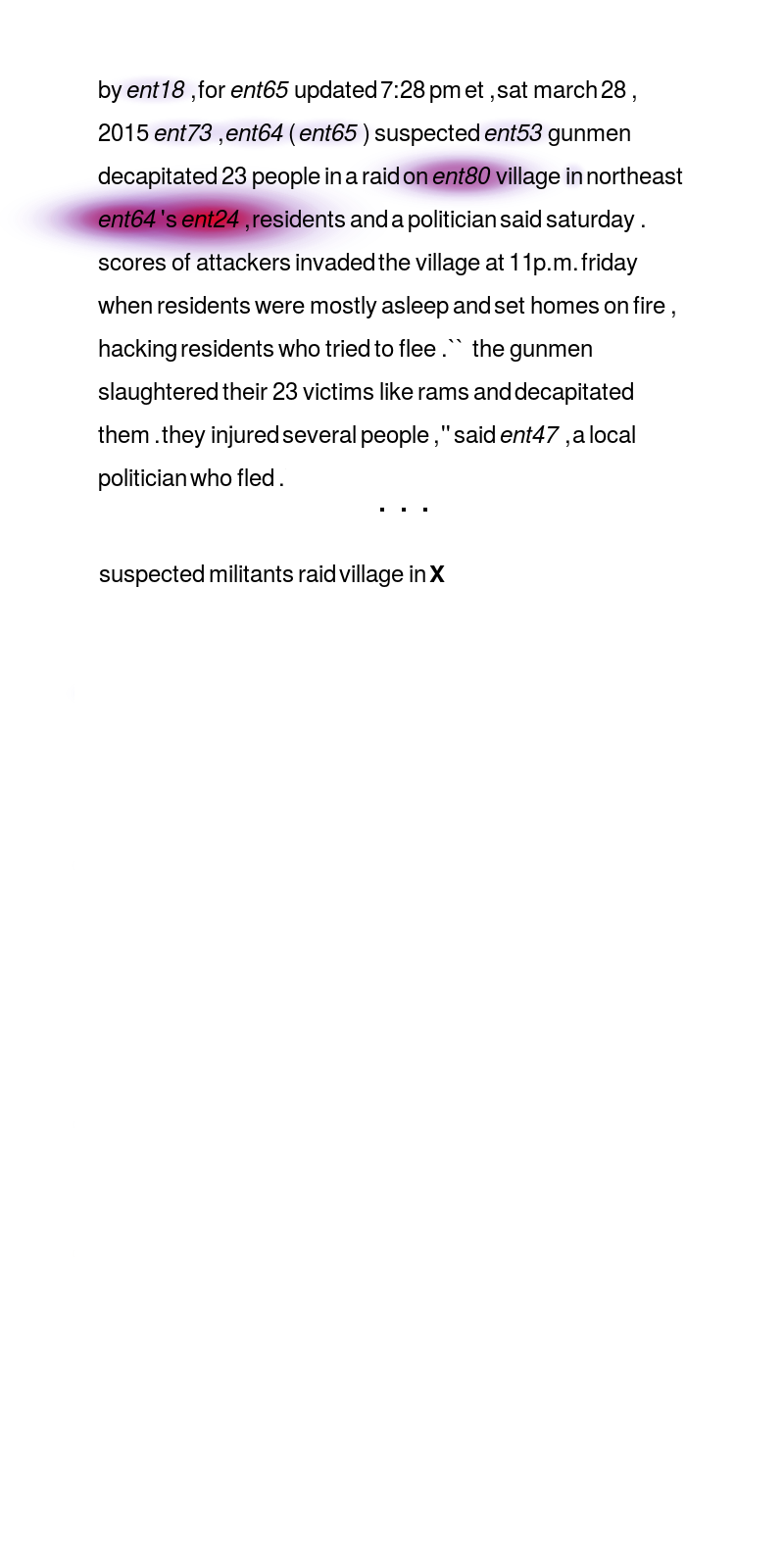}
  \caption{
	   Attention heat maps from the Attentive Reader for two
           wrongly answered validation set queries. In the left case the model
           returns \textit{ent85} (correct: \textit{ent67}), in the right example
           it gives \textit{ent24} (correct: \textit{ent64}). In both cases the
           query is unanswerable due to its ambiguous nature and the model
           selects a plausible answer.
       }
  \label{fig:heatmapsC}
\end{figure}

\begin{figure}[t]
  \centering
  \includegraphics[scale=0.3,clip=true,trim=3cm 30cm 2cm 2cm]{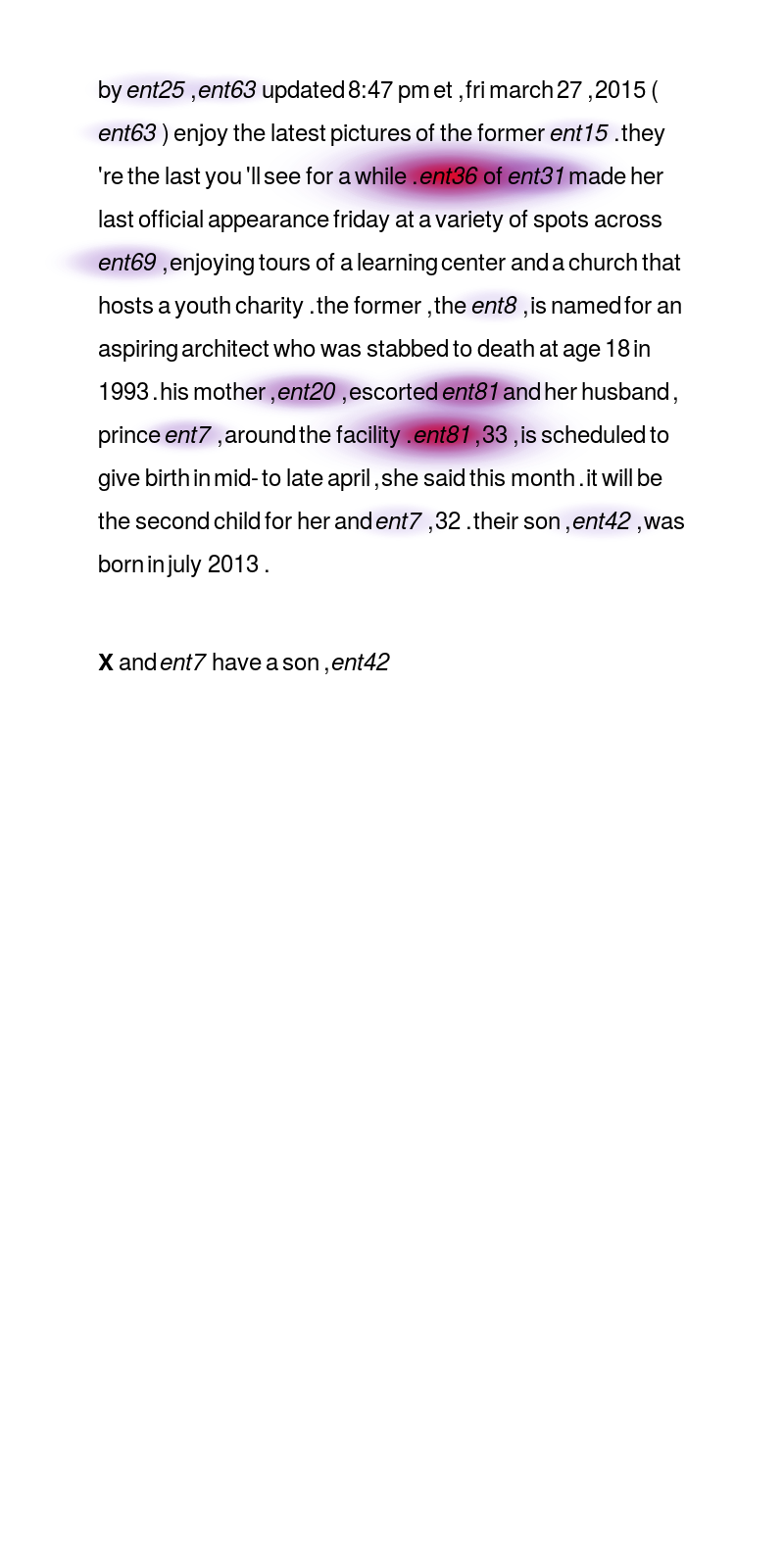}%
  \includegraphics[scale=0.3,clip=true,trim=2cm 30cm 3cm 2cm]{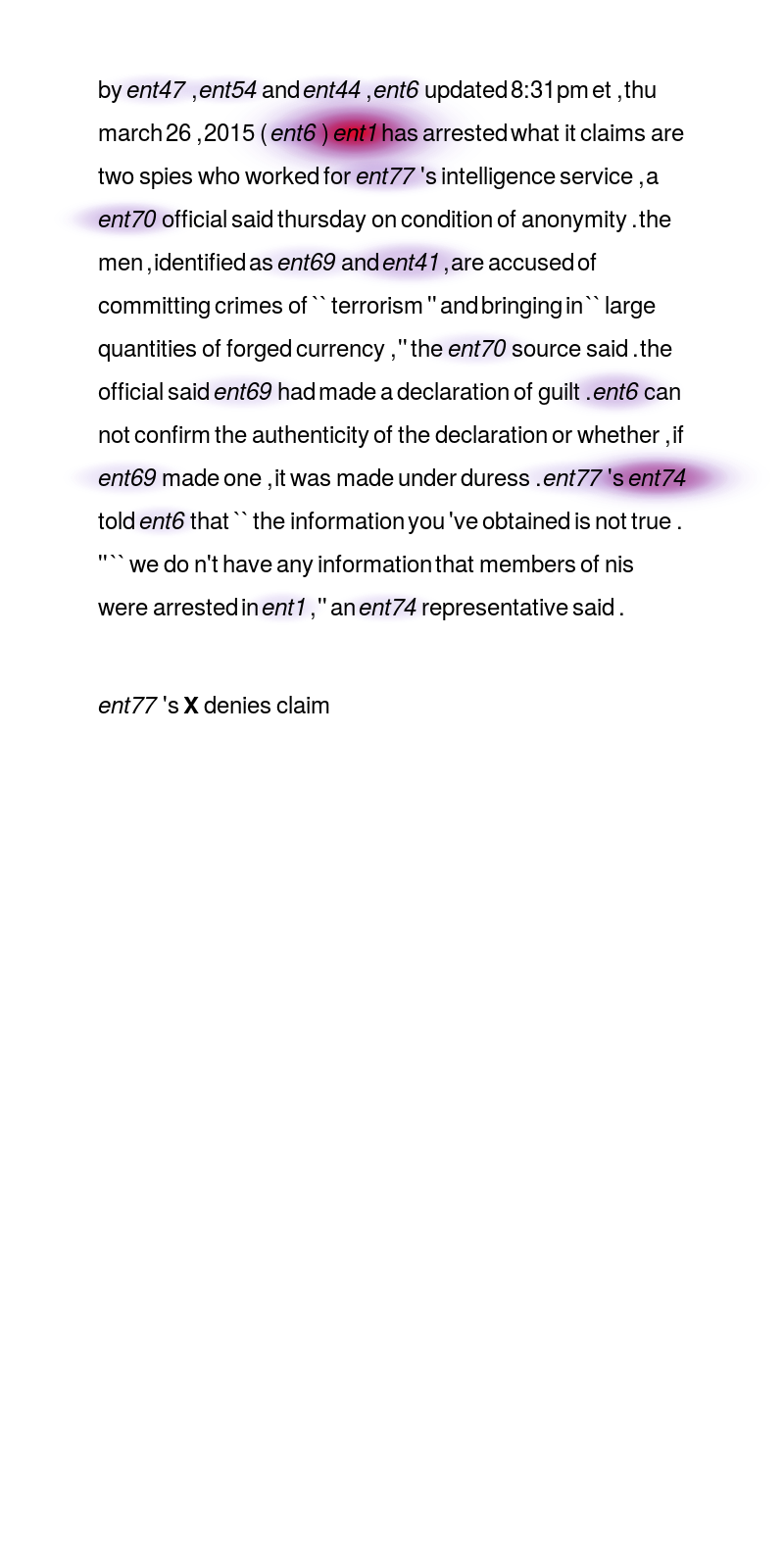}
  \caption{
	   Additional heat maps for negative results. Here the left query
           selected \textit{ent81} instead of \textit{ent15} and the right query
           \textit{ent1} instead of \textit{ent74}.
         }
  \label{fig:heatmapsD}
\end{figure}

\paragraph{Negative Instances}
Figures \ref{fig:heatmapsC} and \ref{fig:heatmapsD} show examples of queries
where the Attentive Reader fails to select the correct answer. The two examples
in Figure \ref{fig:heatmapsC} highlight a fairly common phenomenon in the data,
namely ambiguous queries, where---at least following the anonymisation
process---multiple entities are plausible answers even when evaluated manually.
Note that in both cases the query searches for an entity marker that describes a
geographic location, preceded by the word ``in''. Here it is unclear whether
the placeholder refers to a part of town, town, region or country.

Figure \ref{fig:heatmapsD} contains two additional negative cases. The first
failure is caused by the co-reference entity selection process. The correct
entity, \textit{ent15}, and the predicted one, \textit{ent81}, both refer to the
same person, but not being clustered together. Arguably this is a difficult
clustering as one entity refers to ``Kate Middleton'' and the other to ``The
Duchess of Cambridge''.
The right example shows a situation in which the model fails as it
perhaps gets too little information from the short query and then selects the
wrong cue with the term ``claims'' near the wrongly identified entity
\textit{ent1} (correct: \textit{ent74}).

\subsection{Impatient Reader}

\begin{figure}[h]
  \centering
  \setlength{\fboxsep}{0pt}

  \fbox{\includegraphics[scale=0.22,clip=true,trim=3cm 27cm 3cm 2.5cm]{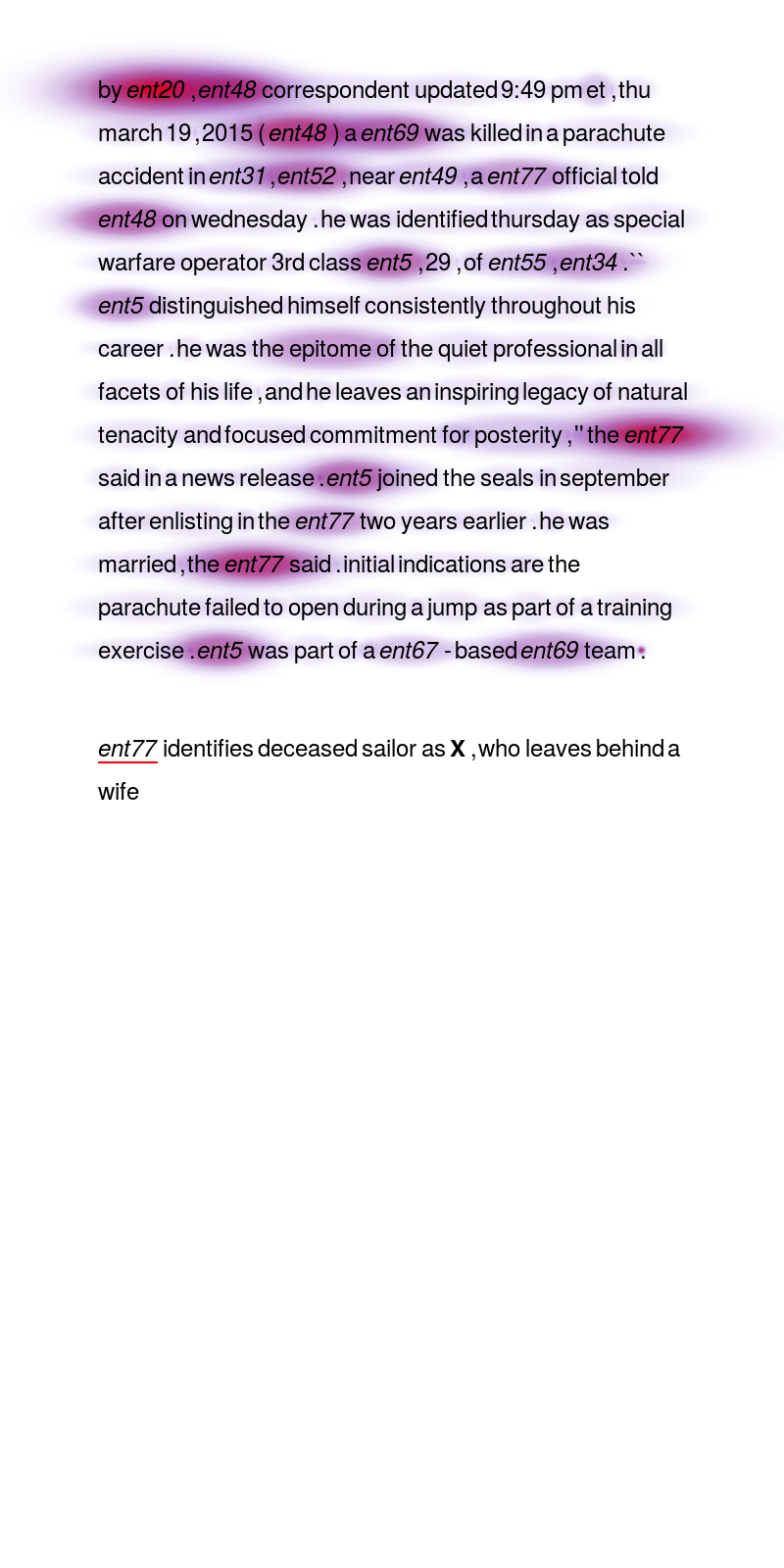}}%
  \fbox{\includegraphics[scale=0.22,clip=true,trim=3cm 27cm 3cm 2.5cm]{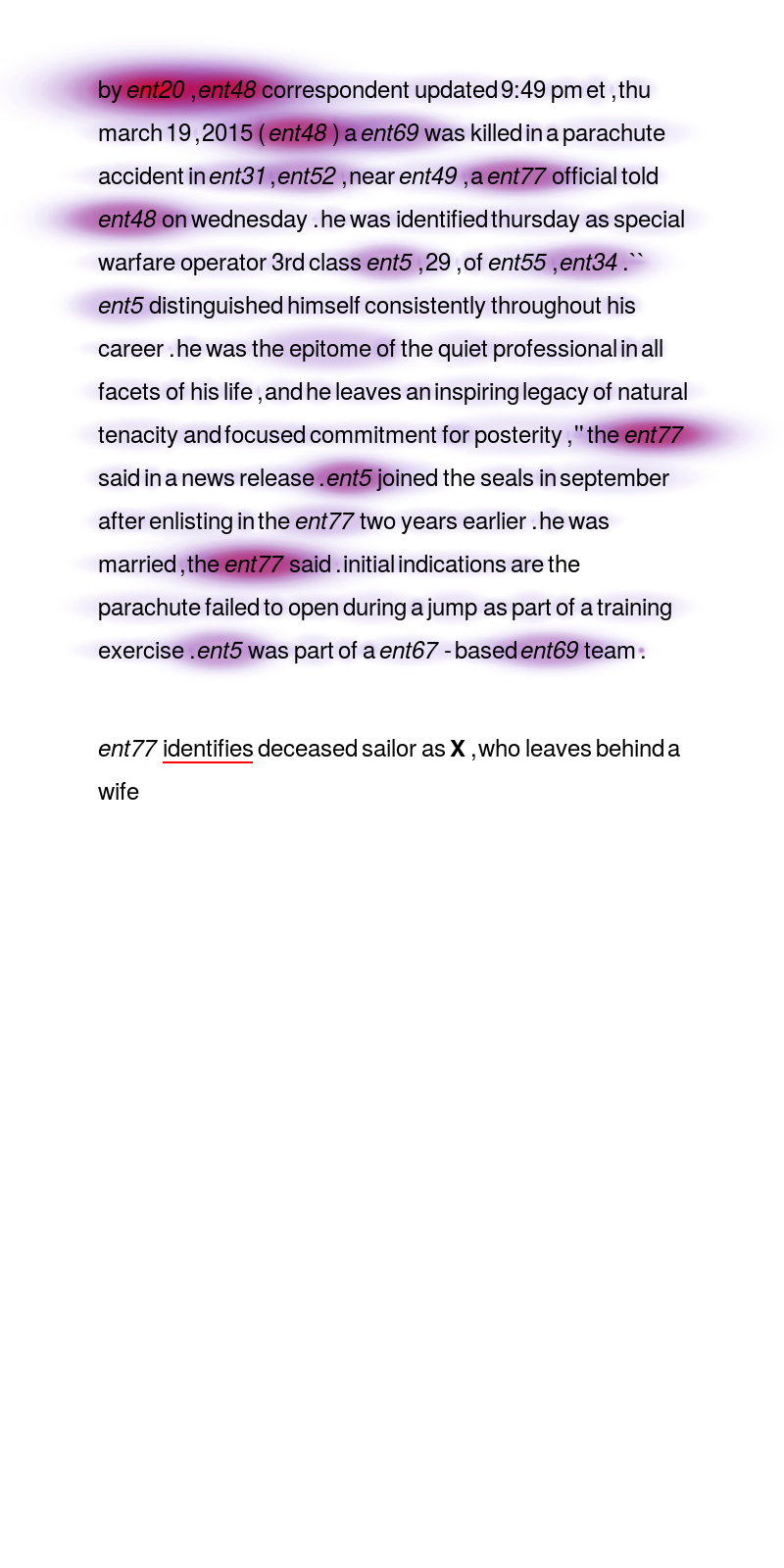}}%
  \fbox{\includegraphics[scale=0.22,clip=true,trim=3cm 27cm 3cm 2.5cm]{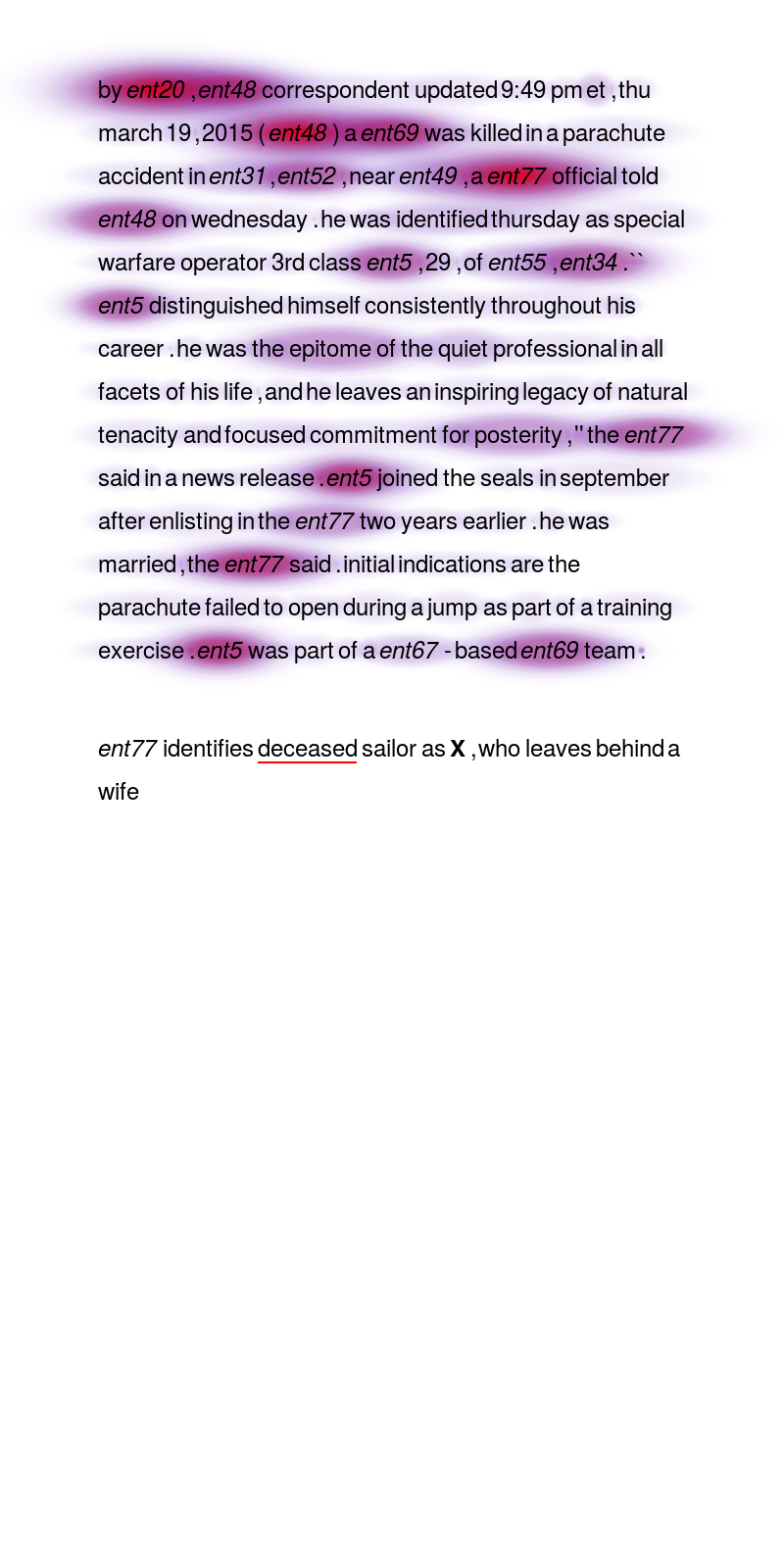}}
  \caption{Attention of the Impatient Reader at time steps 1, 2 and 3.}
  \label{fig:heatmapsE}
\end{figure}

\begin{figure}[h]
  \centering
  \setlength{\fboxsep}{0pt}

  \fbox{\includegraphics[scale=0.22,clip=true,trim=3cm 27cm 3cm 2.5cm]{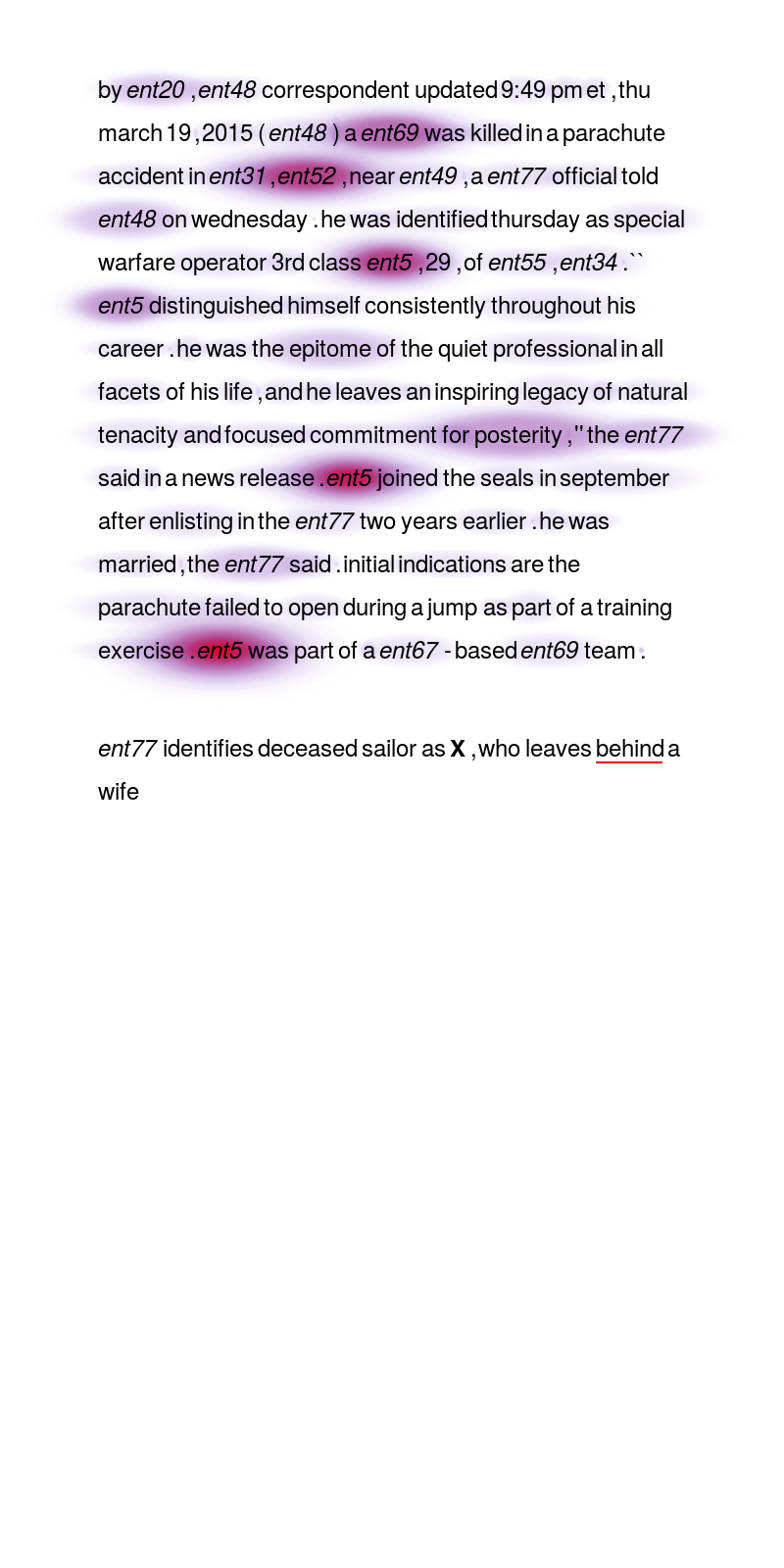}}%
  \fbox{\includegraphics[scale=0.22,clip=true,trim=3cm 27cm 3cm 2.5cm]{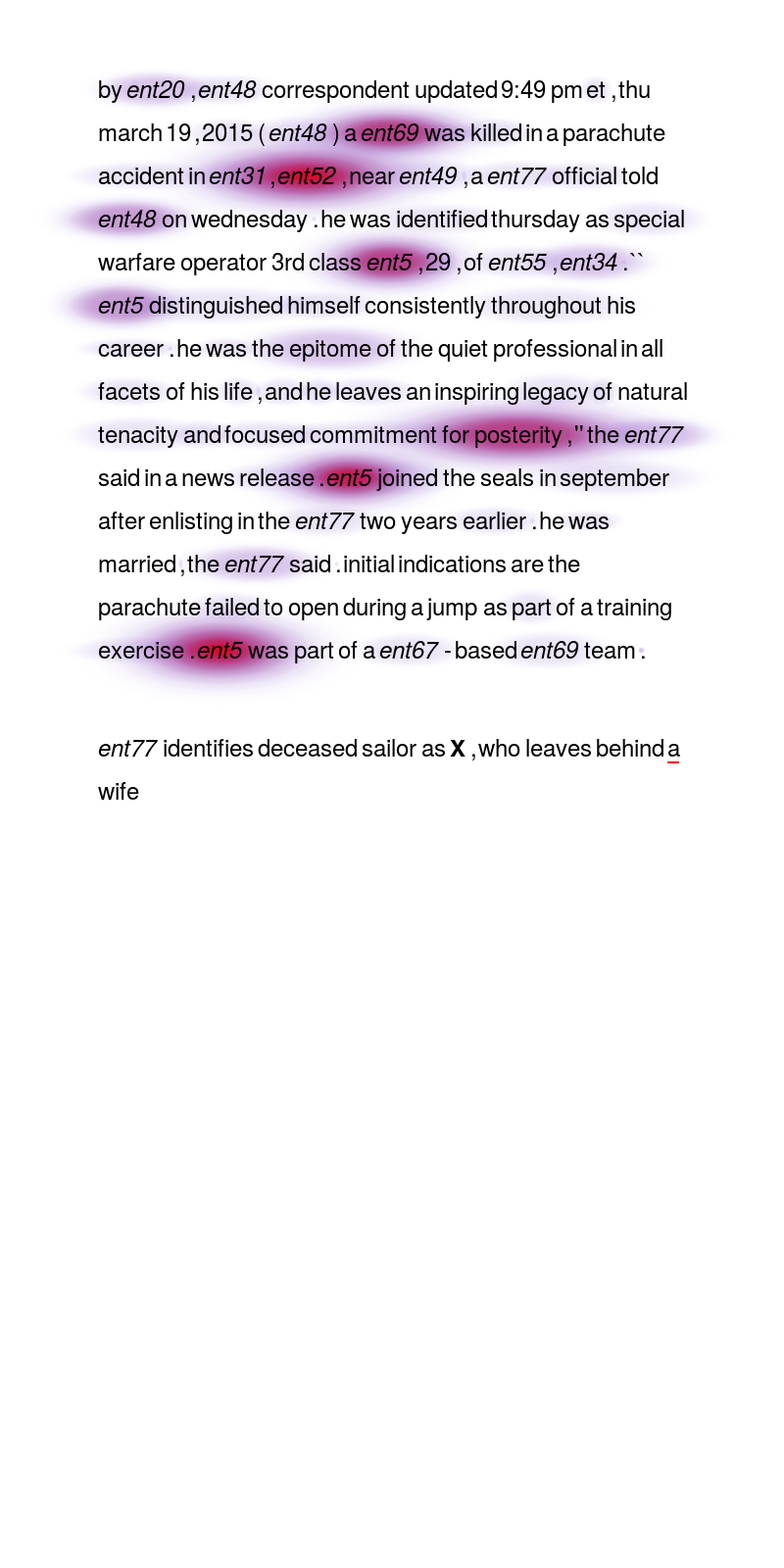}}%
  \fbox{\includegraphics[scale=0.22,clip=true,trim=3cm 27cm 3cm 2.5cm]{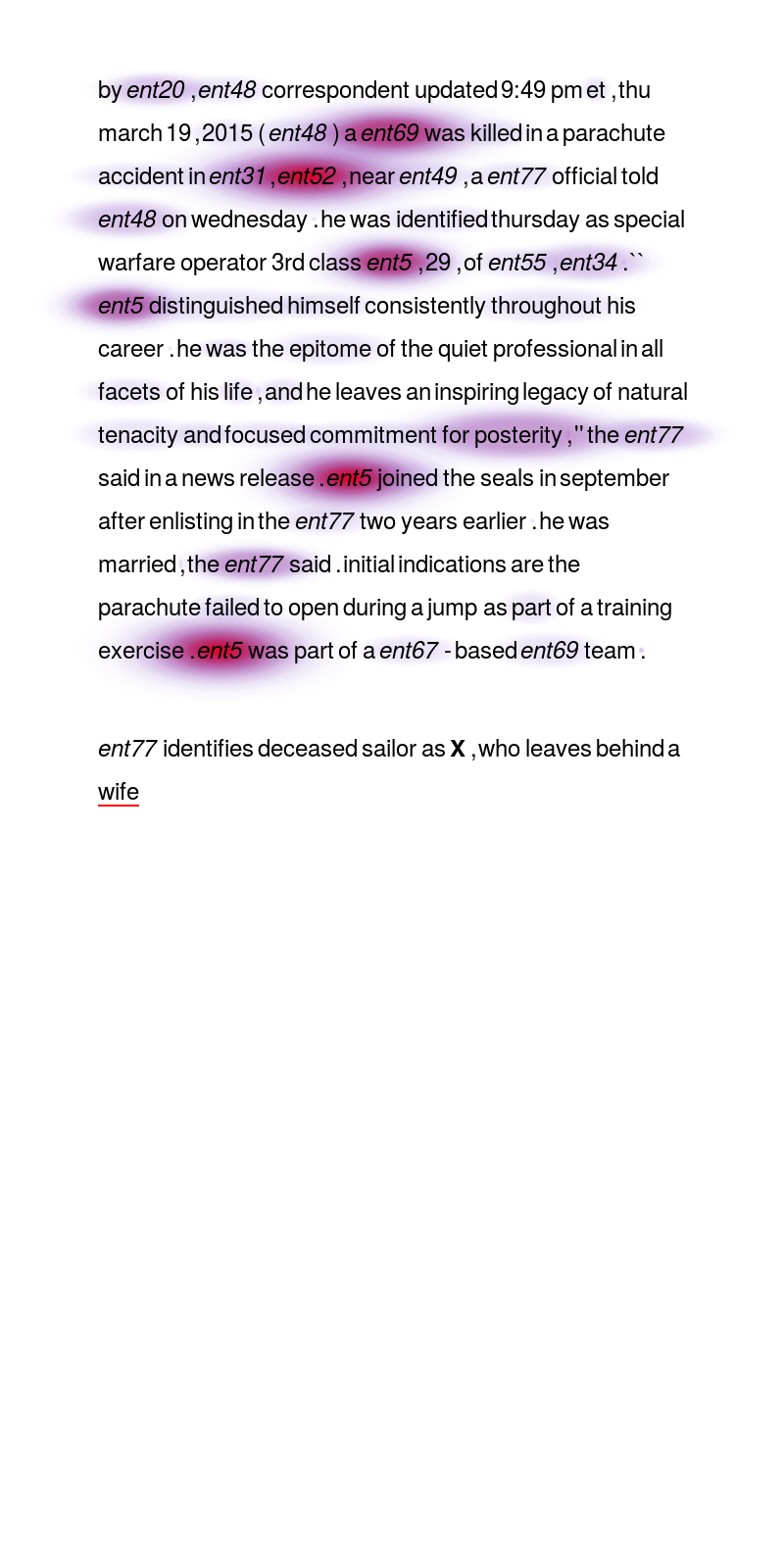}}
  \caption{Attention of the Impatient Reader at time steps 10, 11 and 12.}
  \label{fig:heatmapsZ}
\end{figure}

To give a better intuition for the behaviour of the Impatient Reader, we use a
similar visualisation technique as before. However, this time around we
highlight the attention at every time step as the model updates its focus while
moving through a given query. Figures \ref{fig:heatmapsE}--\ref{fig:heatmapsZ} shows how the attention
of the Impatient Reader changes and becomes increasingly more accurate as the
model considers larger parts of the query.
Note how the attention is distributed
fairly arbitraty at first, slowly focussing on the correct entity
\textit{ent5} only once the question has sufficiently been parsed.

\begin{figure}[h]
  \centering
  \setlength{\fboxsep}{0pt}

  \fbox{\includegraphics[scale=0.22,clip=true,trim=3cm 27cm 3cm 2.5cm]{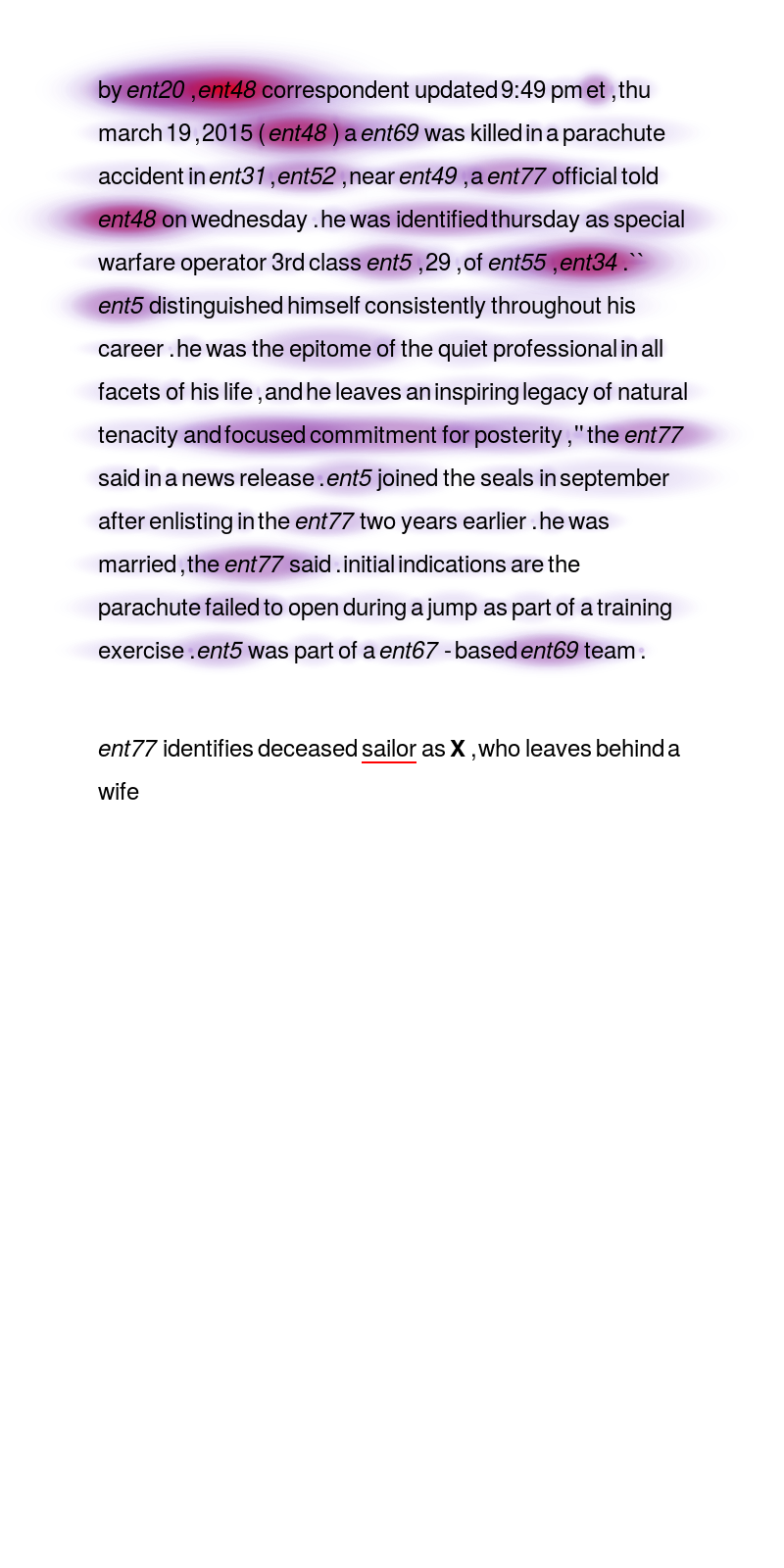}}%
  \fbox{\includegraphics[scale=0.22,clip=true,trim=3cm 27cm 3cm 2.5cm]{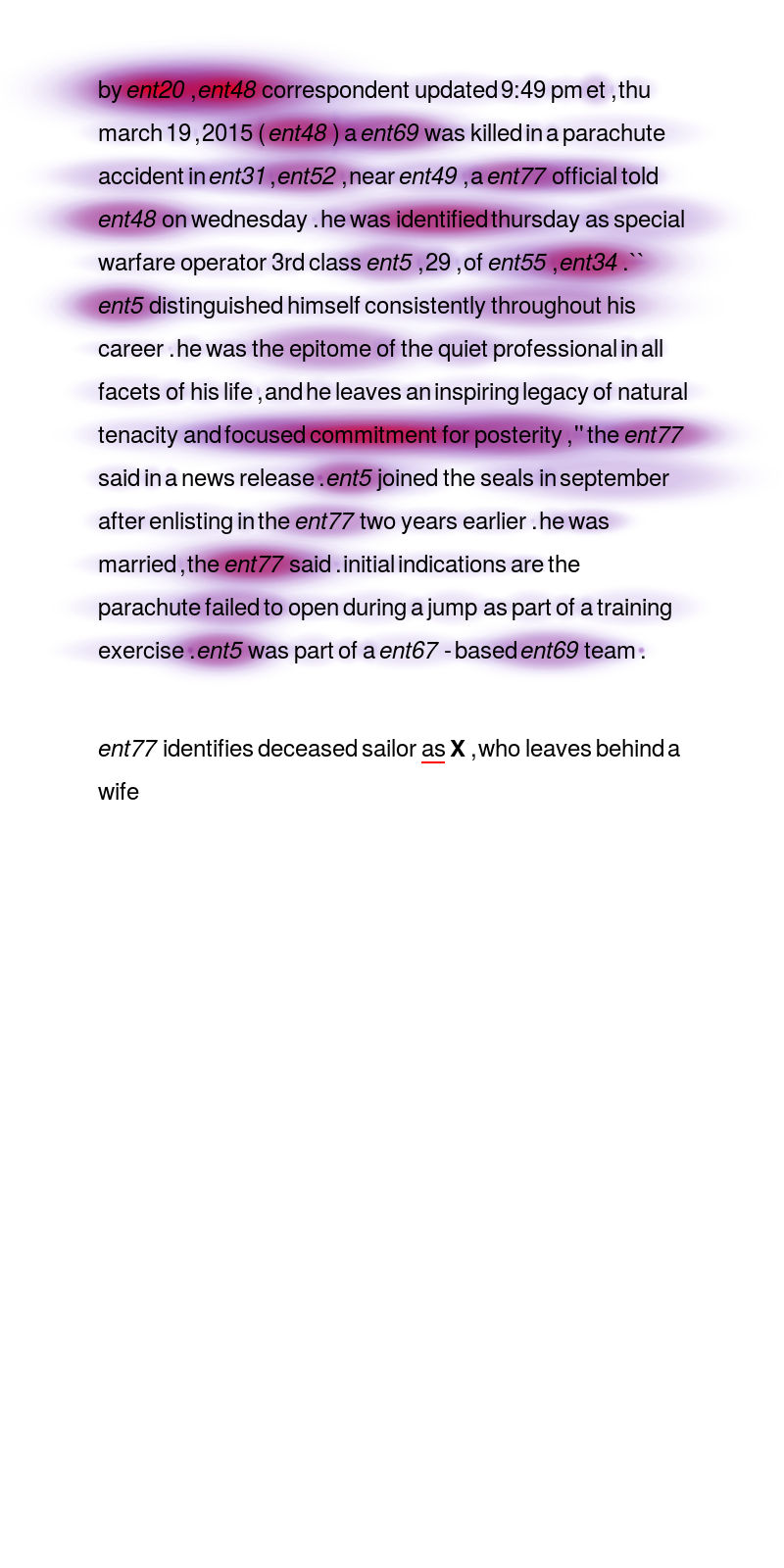}}%
  \fbox{\includegraphics[scale=0.22,clip=true,trim=3cm 27cm 3cm 2.5cm]{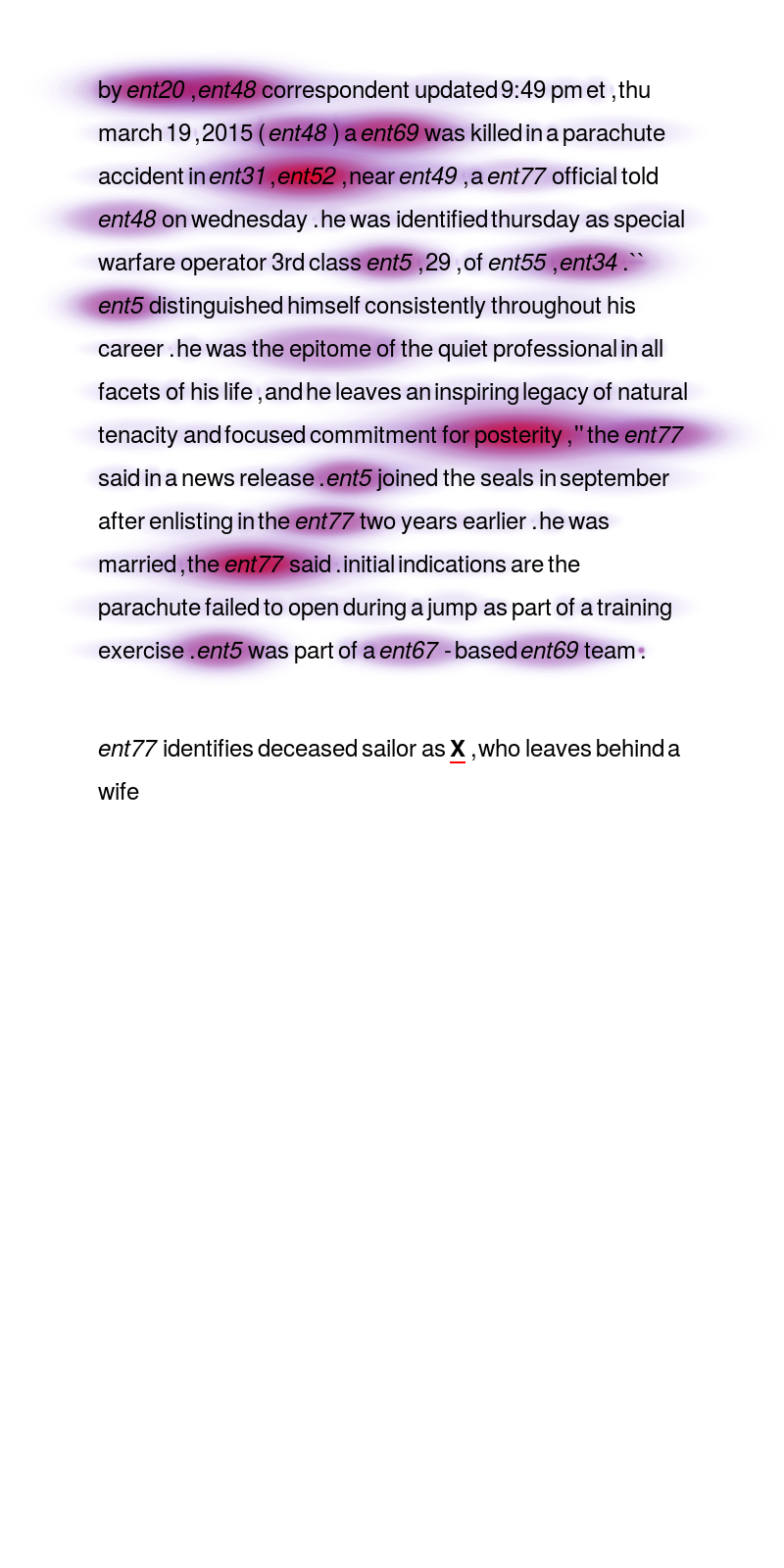}}
  \caption{Attention of the Impatient Reader at time steps 4, 5 and 6.}
\end{figure}

\begin{figure}[h]
  \centering
  \setlength{\fboxsep}{0pt}

  \fbox{\includegraphics[scale=0.22,clip=true,trim=3cm 27cm 3cm 2.5cm]{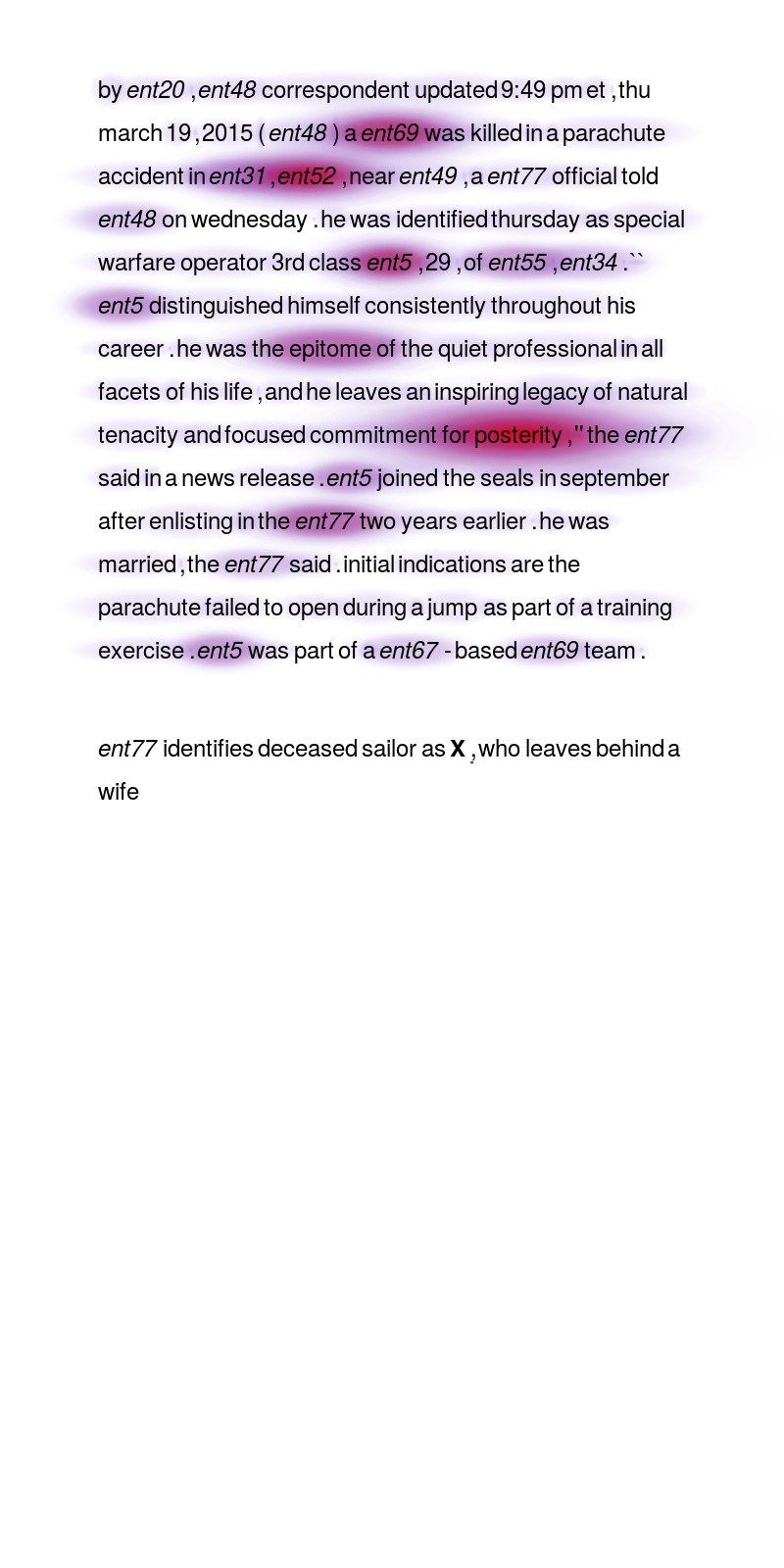}}%
  \fbox{\includegraphics[scale=0.22,clip=true,trim=3cm 27cm 3cm 2.5cm]{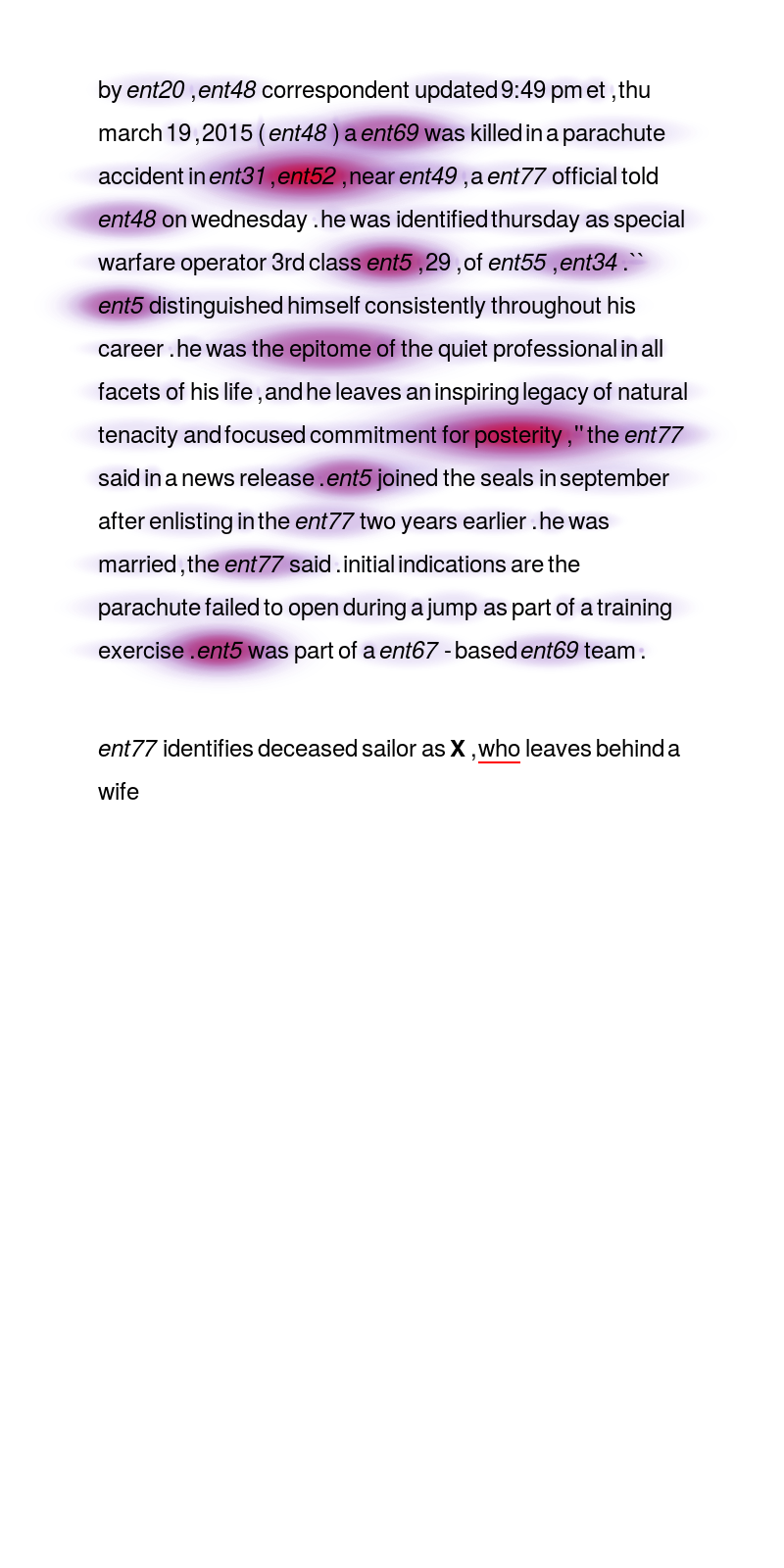}}%
  \fbox{\includegraphics[scale=0.22,clip=true,trim=3cm 27cm 3cm 2.5cm]{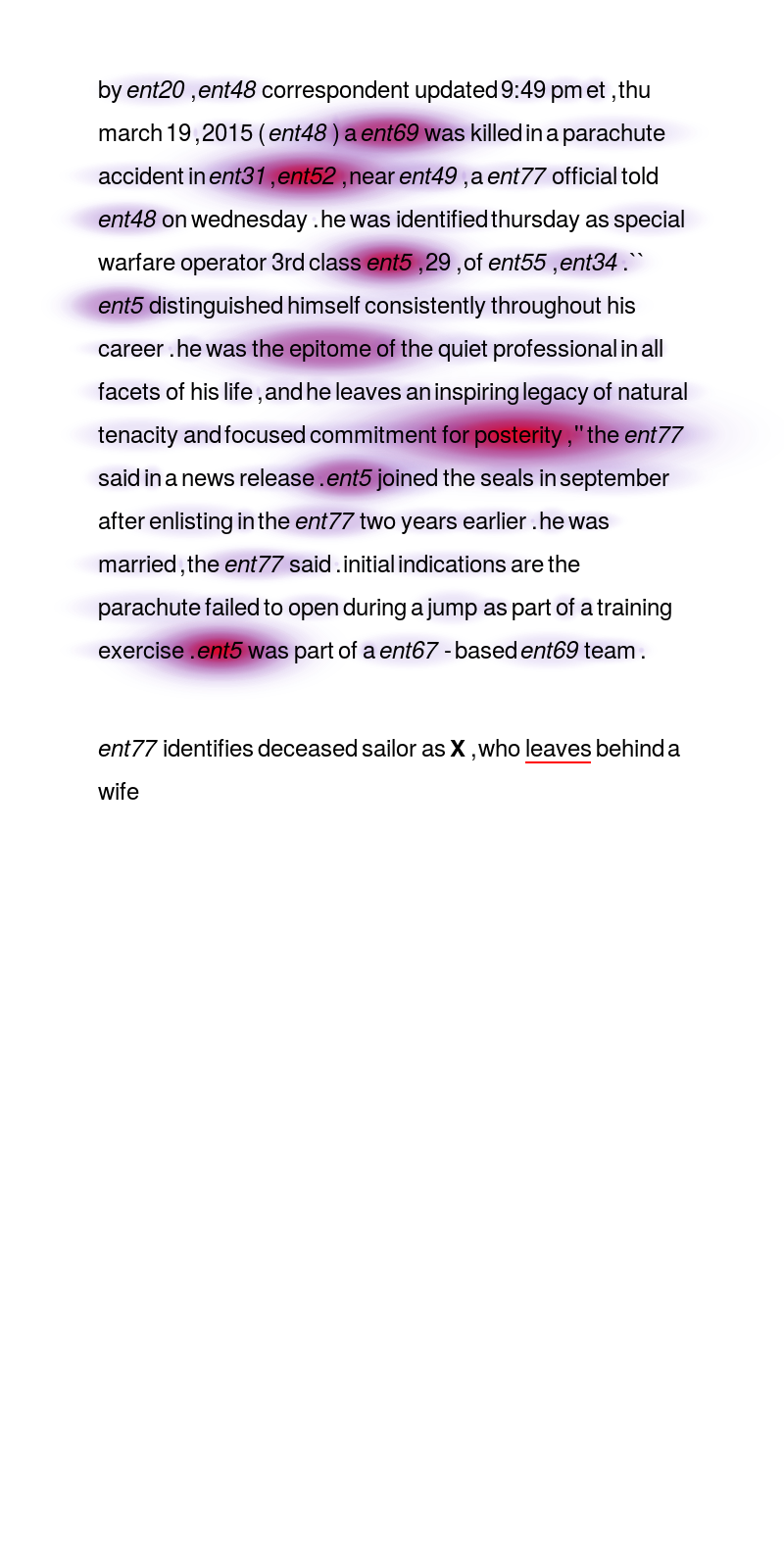}}
  \caption{Attention of the Impatient Reader at time steps 7, 8 and 9.}
\end{figure}

\end{document}